\documentclass{article}

\usepackage[eandd, preprint]{neurips_2026}

\usepackage[utf8]{inputenc} 
\usepackage[T1]{fontenc}    
\usepackage{hyperref}       
\usepackage{url}            
\usepackage{booktabs}       
\usepackage{amsfonts}       
\usepackage{nicefrac}       
\usepackage{microtype}      
\usepackage{xcolor}         
\usepackage{graphicx}
\usepackage{subcaption}
\usepackage{fontawesome5}
\usepackage{etoc}
\usepackage{float}
\usepackage{booktabs}
\usepackage[percent]{overpic}

\usepackage{pifont}
\usepackage{enumitem}
\usepackage{xcolor}
\newcommand{\cmark}{\textcolor{green!60!black}{\ding{51}}} 
\newcommand{\xmark}{\textcolor{red!70!black}{\ding{55}}}   
\usepackage{tabularx}
\usepackage{colortbl}
\usepackage{amssymb}
\usepackage[most]{tcolorbox}
\usepackage{amsmath}
\usepackage{CJKutf8}
\usepackage{makecell}
\usepackage{array}
\usepackage{float}
\usepackage{needspace}
\usepackage{multirow}
\usepackage[table]{xcolor}
\usepackage{bbm}
\usepackage{dsfont}
\usepackage{bm}
\definecolor{elmaBlue}{RGB}{220,235,255}

\title{ELMA: Benchmarking Anaphoric Compositions for Long-Term Text-to-Motion Generation}

\author{%
  {\bfseries
    Shun Kato$^{1,2}$,
    Dixin Yang$^{1}$,
    Kotaro Amaya$^{1}$,
    Yuiga Wada$^{1}$,
  }\\
  {\bfseries
    Kazumi Fukuda$^{2,3}$,
    Takashi Shibuya$^{3}$,
    Yuki Mitsufuji$^{3}$,
    Mariko Isogawa$^{1,2}$
  }\\[1.5ex]
  {\normalfont
    $^{1}$Keio University \quad
    $^{2}$Keio AI Center \quad
    $^{3}$Sony Group Corporation
  }
}

\begin{document}
\maketitle

\begin{center}
  \url{https://elma-dataset.github.io/}
\end{center}

\begin{abstract}
  Text-guided long-term human motion generation is essential for applications such as continuously operating digital avatars. Despite recent progress on short clips, existing models struggle with multi-segment instructions that involve anaphora, where constraints established in earlier segments must be preserved over time. In this paper, we introduce a novel task of long-term motion generation from such anaphoric compositions. To address this gap, we present ELMA, an episodic long-term motion dataset with dense segment-level annotations that preserve cross-segment dependencies. Built via a fully automated collection and filtering pipeline, ELMA extracts high-quality, continuous 30-second clips from in-the-wild videos, capturing the episodic narratives necessary for long-range state modeling. 
  Using ELMA, we benchmark representative long-term motion generation models under independent and anaphoric instruction settings, and study whether contextual fine-tuning improves long-range state modeling. Our extensive experiments provide insights into the capabilities and limitations of existing models in handling context-dependent instructions. These findings highlight the challenge of modeling cross-segment dependencies and offer valuable directions for developing more coherent and semantically consistent long-term motion generation models.
\end{abstract}

\section{Introduction}
\label{sec:intro}
\begin{figure}[ht]
    \centering
    \includegraphics[width=0.8\linewidth]{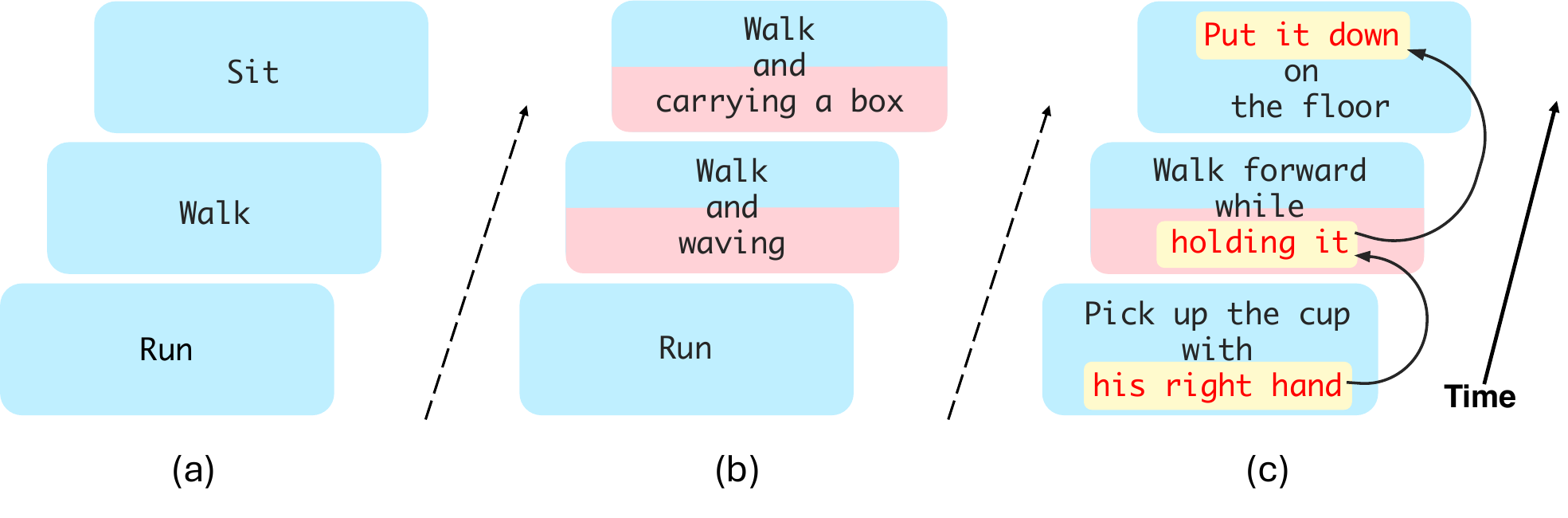}
    \vspace{-0.4em}
    \caption{\textbf{Three types of text composition in long-term motion generation}: As shown in (a), \textit{atomic composition} is the simplest input format, where a sequence is constructed from consecutive segments that each represent a single action. The \textit{concurrent composition} illustrated in (b) is a more complex format than (a), as individual segments may contain two or more actions. Finally, the \textit{anaphoric composition} depicted in (c) is an even more complex format, characterized by cross-dependencies across the input segments.}
    \vspace{-0.8em}
    \label{fig:text_composition}
\end{figure}

\begin{figure}[ht]
    \centering

    \begin{subfigure}[t]{0.48\linewidth}
        \centering
        \includegraphics[width=\linewidth]{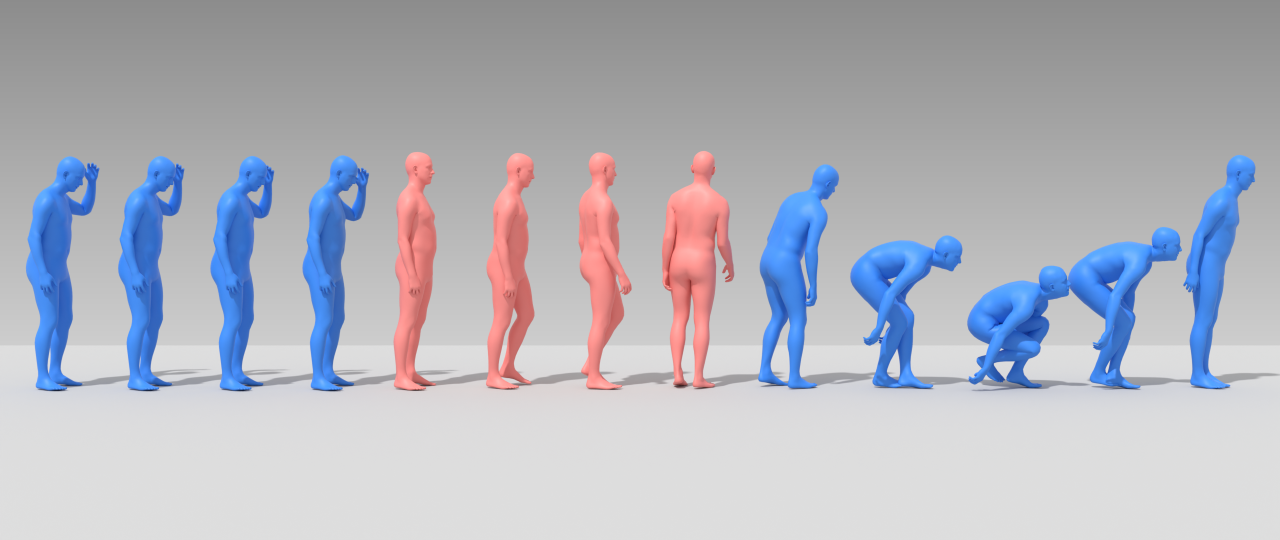}
        \vspace{-0.3em}
        \caption{FlowMDM}
        \label{fig:flowmdm_failure}
    \end{subfigure}
    \hfill
    \begin{subfigure}[t]{0.48\linewidth}
        \centering
        \includegraphics[width=\linewidth]{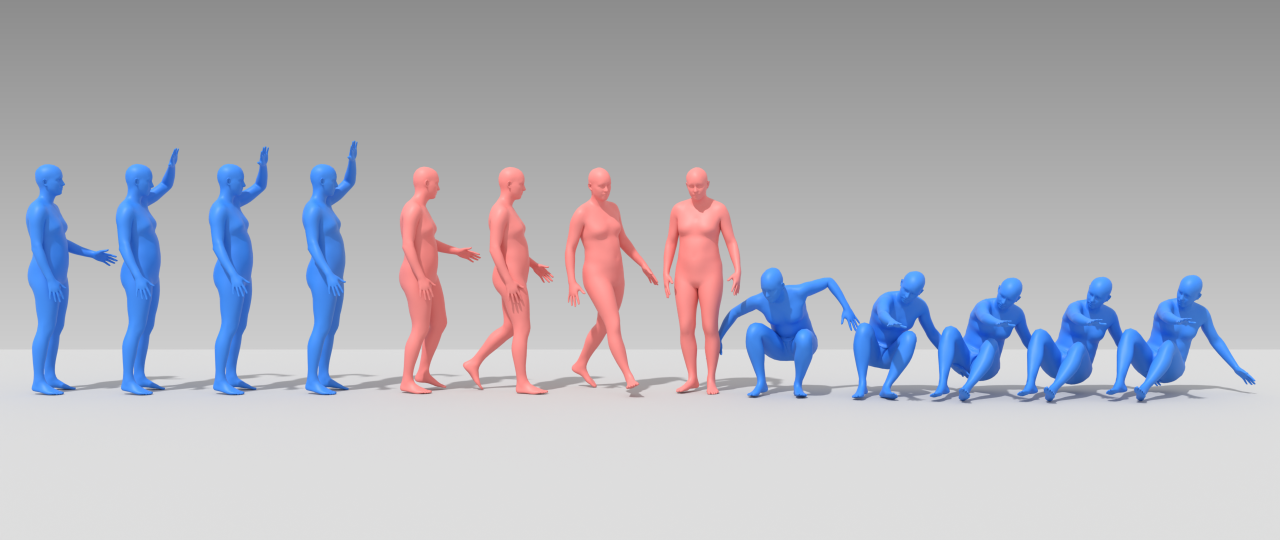}
        \vspace{-0.3em}
        \caption{DoubleTake}
        \label{fig:dart_failure}
    \end{subfigure}

    \vspace{-0.4em}

    \caption{
\textbf{Failure Cases in Anaphoric Motion Generation.}
The input consists of three segment instructions: ``raise the left hand to salute,'' ``walk while saluting,'' and ``sit down.''
Both FlowMDM~\citep{flowmdm} and DoubleTake~\citep{doubletake} fail to follow the second instruction, ``walk while saluting.''
Since existing models treat each segment as an independent input, the second segment cannot inherit the context from the first segment, in which the person is saluting with the left hand.
As a result, the generated motions fail to maintain the left-hand salute while walking.
}
    \vspace{-0.8em}
    \label{fig:failure_case}
\end{figure}

Text-guided human motion generation~\citep{t2m-gpt, mdm, momask, motiondiffuse} aims to synthesize human motion from natural-language descriptions, reducing the need for costly motion capture and manual editing. This technology enables scalable applications in video games, virtual and augmented reality experiences, and robotic systems. In particular, applications that require interactive or continuous control of virtual characters demand effective long-term motion generation methods.

In long-term motion generation, synthesizing arbitrary-length sequences in a single pass is challenging due to computational and precision constraints. As a result, many methods ~\citep{doubletake, stmc, flowmdm, t2lm, unimotion, dartcontrol, motionstreamer, transphase} adopt multi-segment instructions, where a long motion description is specified as a sequence of short segment-level instructions; each instruction guides one motion segment, and segments are concatenated into a full sequence. Existing methods, such as FlowMDM~\citep{flowmdm} and DoubleTake~\citep{doubletake}, have shown promising results in motion quality and text alignment under segment-wise generation; however, the preservation of dependency-induced states across segments remains underexplored. Fig.~\ref{fig:text_composition} illustrates three composition types for multi-segment instructions: atomic, concurrent, and anaphoric. In this work, we focus on anaphoric composition, where later segments refer to entities or constraints introduced earlier (e.g., ``maintaining a salute while walking''). This setting requires explicit state persistence across segments, yet most existing long-term motion datasets and benchmarks annotate and evaluate segments largely in isolation, and therefore do not directly measure this capability. As shown in Fig.~\ref{fig:failure_case}, existing models struggle with long-term motion generation when later segment instructions depend on earlier context.

We therefore propose to treat anaphoric composition as a distinct task setting in long-term motion generation.
To enable this task setting, we identify two key bottlenecks: (i) the lack of episodic long-term datasets with dense, dependency-rich segment captions, and (ii) the inherent limitation of current segment-based generation frameworks, which are typically trained on independent segments and lack the capacity for explicit state propagation.

\noindent
To address the first issue, we introduce ELMA (\textbf{E}pisodic \textbf{L}ong-term \textbf{M}otion dataset with multi-segment \textbf{A}nnotations), a 3D human motion dataset paired with multi-segment text instructions. This dataset comprises training and evaluation sets collected from online RGB video sources using an automated collection pipeline. Combined with a dependency-aware captioning pipeline, it provides a foundation for long-term motion generation from diverse text instructions. The online video clips preserve episodic coherence and naturally occurring long-range dependencies, which are essential for modeling long-term motions with cross-segment constraints.

\noindent
To address the second issue and understand the limitations of current architectures in this new setting, we conducted a comprehensive evaluation of representative baseline methods~\citep{doubletake, flowmdm, unimotion}. Our analysis reveals that while existing methods perform adequately under standard, independent text-to-motion settings, they lack the capacity to explicitly propagate states. Consequently, they suffer from severe temporal inconsistencies and performance drops when subjected to anaphoric constraints. To overcome this, we fine-tuned these models using ELMA to handle anaphoric compositions. ELMA provides supervision for learning to preserve dependency-induced constraints and integrate them into subsequent motion segments, helping bridge the gap between independent segment generation and coherent long-term synthesis. 

\noindent
In summary, our contributions are as follows:
\vspace{-0.4em}
\begin{itemize}
    \item[$\bullet$] \textbf{Task:} We formulate anaphoric text-to-motion as long-term motion generation from multi-segment instructions with cross-segment dependencies.
    \item[$\bullet$] \textbf{Dataset:} We introduce \textbf{ELMA}, a dataset of continuous 30-second 3D motions split into six 5-second segments with context-aware captions and dependency labels.
    \item[$\bullet$] \textbf{Benchmark:} We propose \textbf{ELMABench}, which evaluates dependency resolution and state persistence on anaphoric target segments.
    \item[$\bullet$] \textbf{Baselines and analysis:} We evaluate representative long-term text-to-motion models under independent and anaphoric settings, showing that fine-tuning on ELMA improves realism and semantic alignment.
\end{itemize}
Our project page is available at \url{https://elma-dataset.github.io/}
\vspace{-0.4em}

\section{Related Work}
\noindent \textbf{Dataset for Long-term Human Motion.}
Long-term motion generation requires datasets with long motion sequences and segment-level captions.
BABEL-TEACH~\citep{babel, teach} provides segment-level annotations, but its action-based labels often lack fine-grained details.
Several datasets have enriched textual descriptions by re-annotating AMASS motions, such as FineMotion~\citep{finemotion} and FrankenMotion~\citep{frankenmotion}.
More recent efforts scale long-term motion data from in-the-wild videos, including HuMo100M~\citep{motionctrl} and OpenT2M~\citep{opent2m}.
However, these datasets either rely on pose-centric descriptions or annotate each segment independently, and thus do not explicitly capture episodic narratives with cross-segment dependencies.
This leaves a critical gap for learning long-term motion generation under anaphoric constraints.

\noindent \textbf{Long-term Motion Generation Models. }
Recent long-term motion generation models extend text-to-motion synthesis to multi-segment instructions through latent blending~\citep{doubletake}, timeline-based control~\citep{flowmdm, stmc}, autoregressive motion priors~\citep{dartcontrol}, and hierarchical control~\citep{unimotion}.
Despite their significant strides in producing visually smooth transitions across sequential instructions, these models fundamentally struggle with anaphoric compositions. This critical limitation primarily arises from the absence of suitable training datasets featuring episodic, dependency-rich multi-segment annotations. Without large-scale data explicitly demonstrating how to preserve and propagate established states across long temporal horizons, current state-of-the-art models cannot learn to maintain temporal consistency under anaphoric constraints. This unresolved data gap directly motivates the necessity of our proposed ELMA dataset and the accompanying fine-tuning paradigm.

\section{Anaphoric Text to Motion}
\label{sec:task}
We formulate \textit{anaphoric motion generation} as a long-term motion generation task from multi-segment text instructions with cross-segment dependencies.
Given a sequence of $T$ segment-level instructions 
$\mathbf{x}_{1:T}=(\mathbf{x}^{(1)},\ldots,\mathbf{x}^{(T)})$
and their segment lengths 
$\boldsymbol{\ell}_{1:T}=(\ell_1,\ldots,\ell_T)$, where $\sum_{t=1}^{T}\ell_t=N$, the goal is to generate a motion sequence
$\mathbf{m}_{1:N}\in\mathbb{R}^{N\times D}$, where $N$ is the total number of frames and $D$ is the motion feature dimension.
Unlike standard segment-wise text-to-motion settings, the interpretation of a target instruction $\mathbf{x}^{(t)}$ may depend on the preceding context $\mathbf{x}_{1:t-1}$.
Thus, the generated motion must be locally consistent with each segment instruction and duration, while also resolving cross-segment dependencies and preserving previously established actions, object states, and body-part constraints throughout the sequence. Formally, let $p_{\mathrm{data}}(\mathbf{m}_{1:N}\mid \mathbf{x}_{1:T},\boldsymbol{\ell}_{1:T})$ denote the conditional distribution of valid long-term motions given the multi-segment instruction sequence and segment lengths.
The task is to learn a generative model $p_\theta$ that approximates this distribution, generating motions as 
$\hat{\mathbf{m}}_{1:N}\sim p_\theta(\mathbf{m}_{1:N}\mid \mathbf{x}_{1:T},\boldsymbol{\ell}_{1:T})$.
We instantiate this task in ELMABench (Sec.~\ref{sec:bench}).
\vspace{-0.4em}
\section{ELMA Dataset}
\subsection{Overview}
\begin{figure}[H]
    \centering
    \vspace{-0.5em}
    \includegraphics[width=1.0\linewidth]{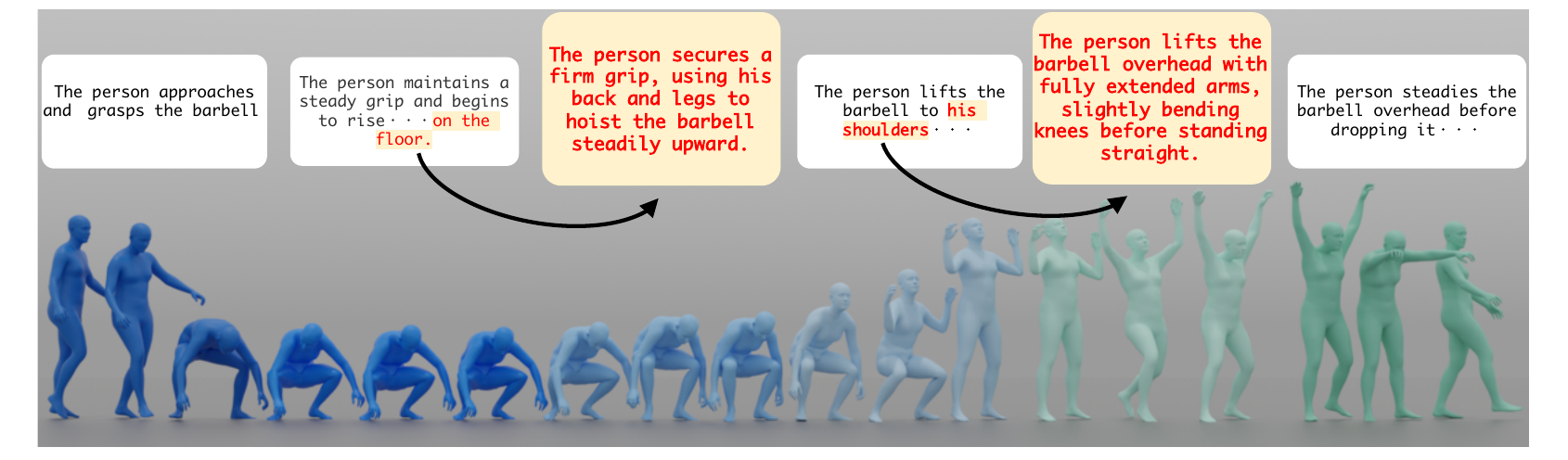}
    \vspace{-0.5em}
    \caption{
    \textbf{Example from the ELMA dataset.} Highlighted phrases indicate cross-segment dependencies between segment captions, with arrows pointing from source segments to their dependent segments. For example, the fifth segment, \textit{“The person lifts the barbell overhead with fully extended arms, slightly bending knees before standing straight”}, is clarified by the earlier phrase \textit{“to his shoulders”}, which indicates that the person lifts the barbell overhead from shoulder level. Such dependencies can occur not only between adjacent segments but also between non-adjacent segments. Therefore, to generate the correct motion from these multi-segment inputs, a model must resolve dependencies across segments.
    }
    \label{fig:data_analysis}
    \vspace{-0.7em}
\end{figure}
Learning and evaluating Anaphoric Text to Motion (Sec.~\ref{sec:task}) requires datasets in which later segment descriptions can depend on earlier context.
However, existing long-term motion datasets~\citep{babel, teach, motionctrl, opent2m, dmc, humanml3d, frankenmotion, finemotion} either stitch together independent motion clips~\citep{motionctrl, dmc}, annotate each segment with isolated text descriptions~\citep{opent2m, finemotion}, or compress detailed motion descriptions into action labels~\citep{babel, teach}, thereby breaking semantic continuity across segments and removing the contextual cues needed to express such dependencies.
We therefore introduce ELMA (\textbf{E}pisodic \textbf{L}ong-term \textbf{M}otion dataset with multi-segment \textbf{A}nnotations), an episodic long-term 3D human motion dataset built from continuous 30-second clips with densely aligned segment-level captions that explicitly preserve and annotate anaphoric dependencies across segments.
Unlike existing datasets that often lack temporal semantic coherence due to the aggregation of fragmented short clips or the synthesis of motion data using existing generative models, our approach leverages continuous in-the-wild RGB videos. We choose 30 seconds as a practical horizon that is long enough to exhibit cross-segment state persistence while remaining feasible for reliable tracking in unconstrained videos. By extracting uninterrupted 30-second sequences, ELMA naturally preserves the temporal semantic coherence and physical consistency of extended activities. These episodic clips provide the coherent context needed to capture long-range constraints, such as maintaining a hand pose while transitioning to locomotion or sustaining an object-holding state across segments.
\begin{table}[H]
  \centering
  \caption{
  \textbf{Dataset comparison.} ``Motion'' indicates whether samples are built from continuous motion sequences, while ``Caption'' indicates whether segment-level captions are written with preceding context rather than independently.
  }
  \label{tab:dataset_comparison}
  \footnotesize
  \renewcommand{\arraystretch}{1.08}
  \setlength{\tabcolsep}{6pt}
  \resizebox{\linewidth}{!}{%
    \begin{tabular}{lcccccccc}
      \toprule
      \multirow{2}{*}{Dataset}
      & \multirow{2}{*}{Clips}
      & \multirow{2}{*}{Hours}
      & \multirow{2}{*}{Avg. dur.}
      & \multicolumn{2}{c}{Continuity}
      & \multicolumn{3}{c}{Type} \\
      \cmidrule(lr){5-6}
      \cmidrule(lr){7-9}
      & & & & Motion & Caption & Atomic & Concurrent & Anaphoric \\
      \midrule
      BABEL~\citep{babel}
        & 13.2k & 43.5 & 11.9s & \cmark & \xmark & \cmark & \xmark & \xmark \\
      HuMMan-MoGen~\citep{hummanmogen, finemogen}
        & 6.2k & -- & -- & \xmark & \xmark & \cmark & \xmark & \xmark \\
      FineMotion~\citep{finemotion}
        & 29k & 57.18 & 7.04s & \xmark & \xmark & \cmark & \cmark & \xmark \\
      CompMo~\citep{dmc}
        & 60k & 664.67 & 39.9s & \xmark & \xmark & \cmark & \xmark & \xmark \\
      FrankenStein~\citep{frankenmotion}
        & 16.1k & 39.07 & 8.74s & \cmark & \xmark & \cmark & \xmark & \xmark \\
      HuMo100M-I2LM~\citep{motionctrl}
        & 500k+ & -- & -- & \xmark & \xmark & \cmark & \cmark & \xmark \\
      OpenT2M~\citep{opent2m}
        & 1M+ & 2815 & 10.1s & \cmark & \xmark & \cmark & \cmark & \xmark \\
      \midrule
      \rowcolor{elmaBlue}
      \textbf{ELMA}
        & 13.2k & 110 & 30.0s & \cmark & \cmark & \cmark & \cmark & \cmark \\
      \bottomrule
    \end{tabular}%
  }
\end{table}

\subsection{Dataset Construction Pipeline}
\begin{figure}[H]
    \centering
    \vspace{-0.5em}
    \includegraphics[width=1.0\linewidth]{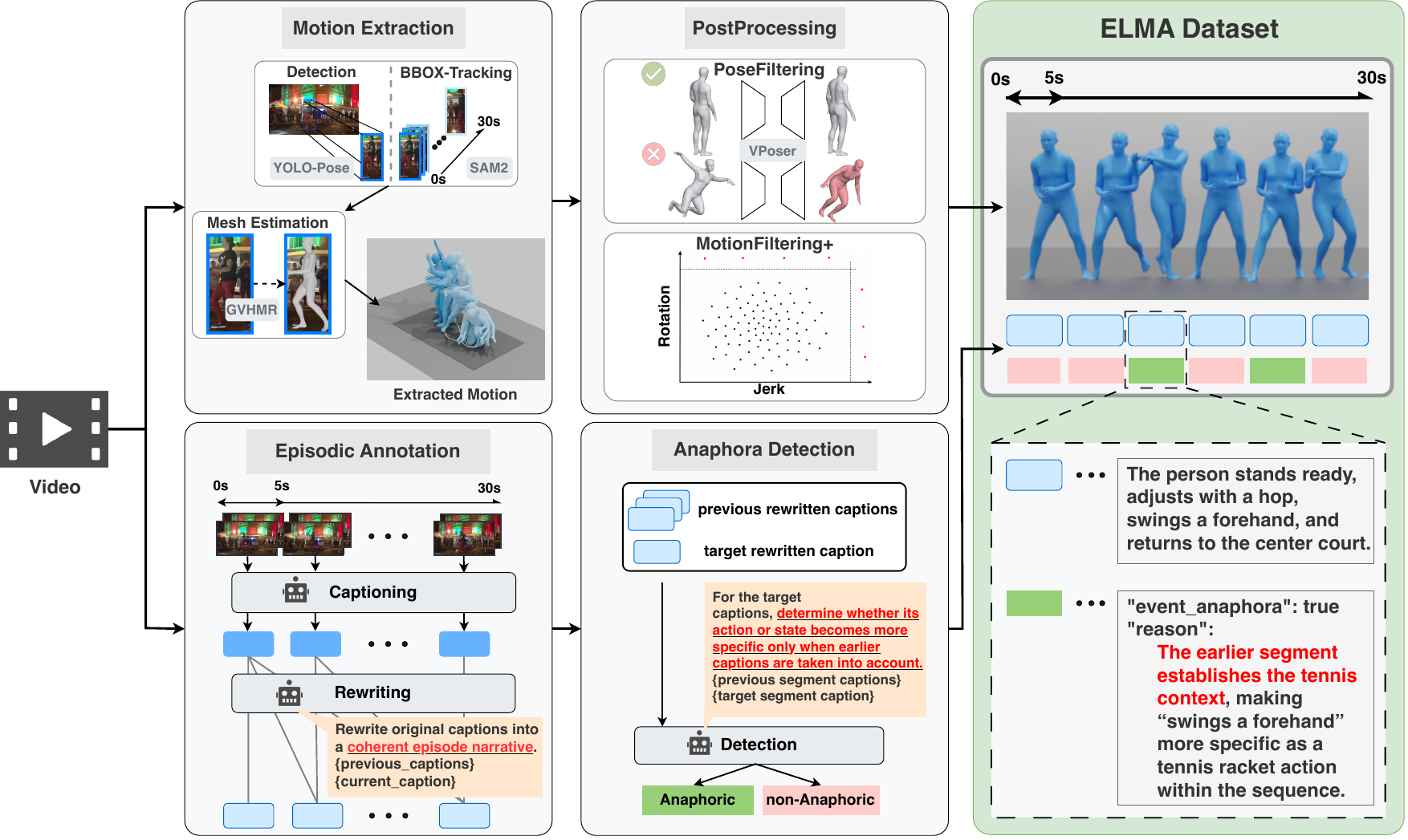}
    \vspace{-0.6em}
    \caption{
    \textbf{Overview of the ELMA dataset construction pipeline.} The upper stream illustrates motion extraction and quality control, while the lower stream illustrates episodic text annotation and anaphora detection. The entire pipeline is fully automated.
    }
    \label{fig:pipeline}
    \vspace{-0.8em}
\end{figure}
To efficiently collect high-quality motion data from these video sources, we developed an automated collection pipeline, as illustrated in Fig.~\ref{fig:pipeline}. The pipeline consists of two streams: (i) motion extraction and quality control, and (ii) episodic text annotation with anaphora detection.

In motion extraction, we initialize person tracking using pose estimation and propagate person masks across frames, followed by two post-processing steps to obtain high-quality 3D motion sequences. In annotation, we first divide each 30-second sequence into six 5-second segments and generate context-aware captions that form a coherent episode. We then detect anaphoric segments and identify their dependencies on preceding segments.

\subsubsection{Motion Extraction}
First, we collect online videos using action-related queries (spanning daily activities, sports, and interactions; details in Appendix~\ref{app:search_queries}) and segment them into scene-level clips with PySceneDetect\footnote{\url{https://github.com/Breakthrough/PySceneDetect}}. 
This process yields more than 5 million scene candidates; however, to ensure that the selected clips sufficiently preserve long-term motion context, we retain only scenes longer than 30 seconds, from which we later extract a 30-second window where a target person remains continuously visible.

However, even scenes retrieved using action-related queries do not always contain human motions suitable for motion extraction. To address this issue, we combine keypoint detection with YOLO-Pose~\citep{yolo-pose} and bounding box tracking with SAM~2~\citep{sam2} to track persons within each scene and extract 30-second clips in which the full body remains continuously visible.  In particular, keypoint-based methods are more effective than conventional BBox-based methods at detecting occlusion, enabling strict full-body visibility assessment based on confidence scores. Details of the tracking and filtering pipeline are provided in Appendix~\ref{app:preprocessing}.

Finally, we apply GVHMR~\citep{gvhmr} to the filtered clips to extract SMPL-X parameters. GVHMR is a model that enables human mesh recovery aligned with the world coordinate system, and it serves as the foundation of the motion representation used in this work. Through this collection and reconstruction pipeline, we ultimately obtain more than 20k motion clips.
\subsubsection{Postprocessing}
\begin{figure}
    \centering

    \begin{subfigure}[t]{0.49\linewidth}
        \centering
        \includegraphics[width=\linewidth]{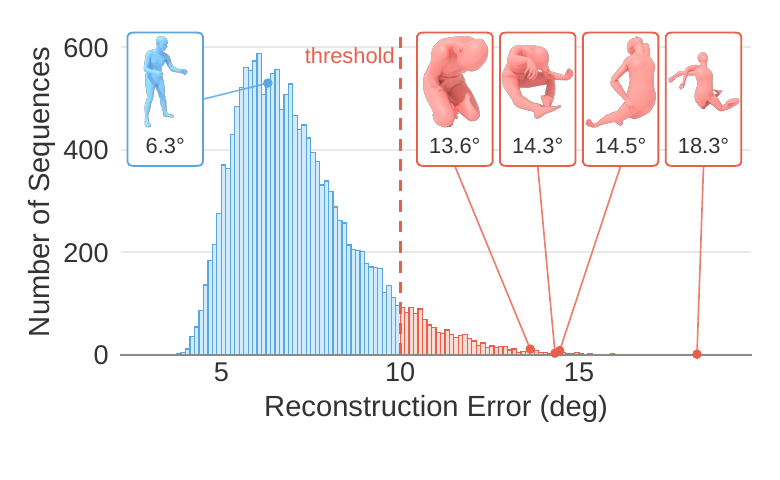}
        \vspace{-8mm}
        \caption{VPoser Reconstruction Error}
        \label{fig:PoseFiltering}
    \end{subfigure}
    \hfill
    \begin{subfigure}[t]{0.49\linewidth}
        \centering
        \includegraphics[width=\linewidth]{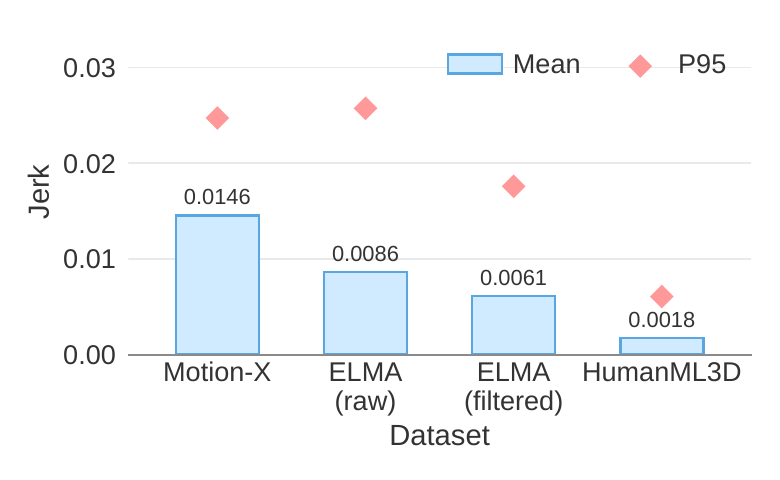}
        \vspace{-8mm}
        \caption{Jerk Statistics Across Motion Datasets}
        \label{fig:MotionFiltering}
    \end{subfigure}
    \caption{
\textbf{Quality control in the ELMA dataset. }
(a) Distribution of VPoser reconstruction errors used for PoseFiltering. The red dashed line denotes the filtering threshold, and poses above this threshold are treated as outliers. Example rejected poses are shown on the right. 
(b) Jerk statistics across motion datasets. ELMA (raw) denotes our reconstructed motions before filtering, while ELMA (filtered) denotes the results after applying MotionFiltering+. Lower jerk indicates smoother temporal motion dynamics.
}
    \vspace{-3mm}
    \label{fig:postprocessing}
\end{figure}
Motion sequences reconstructed from videos through the motion collection pipeline often contain artifacts caused by occlusion, abrupt body motion, and estimation errors. This challenge is amplified in long-duration RGB videos, where even short tracking failures or pose estimation errors can accumulate over time and corrupt an otherwise usable sequence. To ensure the reliability of long-term motion clips, we therefore apply a two-stage filtering procedure using AMASS as a clean reference distribution, aiming to remove clearly corrupted clips while preserving valid motion diversity.

\textbf{PoseFiltering (Stage 1)} removes clips with unnatural poses using a VPoser~\citep{smplx}-based pose naturalness score aggregated at the clip level. \textbf{MotionFiltering+ (Stage 2)} removes clips with abnormal temporal dynamics based on jerk and frame-to-frame joint rotation changes, and additionally discards near-static failure cases. Full definitions, thresholds, and implementation details are provided in Appendix~\ref{app:postprocessing}.
Fig.~\ref{fig:postprocessing} illustrates the effect of our two-stage postprocessing.
Fig.~\ref{fig:PoseFiltering} shows representative pose examples removed by PoseFiltering.
The rejected examples include structurally implausible poses that often arise in challenging in-the-wild videos, such as unstable balancing poses during climbing or grasping motions, as well as inverted poses.
Fig.~\ref{fig:MotionFiltering} analyzes the change in jerk statistics before and after applying MotionFiltering+. The filtered-out samples typically include sequences with temporal jittering, abrupt joint rotations, implausible or extreme poses, as well as clips containing little to no motion.
Despite being reconstructed from long-term unconstrained RGB videos, where occlusion and tracking failures are common, ELMA (filtered) achieves lower mean and P95 jerk than Motion-X~\citep{motion-x}.
After these postprocessing processes, we discard 25\% of the collected motion sequences and retain the remaining 75\% as high-quality motion data. 

\subsubsection{Episodic Text Annotation}
In multi-segment instruction settings, users typically provide the next motion instruction incrementally while taking into account the preceding motion context. To achieve annotations that better reflect this natural human input style than existing long-term datasets, which annotate each segment independently, we introduce Episodic Text Annotation. We use 5-second segments as a practical compromise between fine-grained controllability and reliable captioning in unconstrained videos. Specifically, we first divide each 30-second video into six 5-second segments and generate an independent annotation for each segment using Gemini 2.5 Pro~\citep{gemini25pro}. We then rewrite each segment caption at time $t$ with the same model, conditioning on the preceding captions up to time $t-1$, to encourage natural episodic continuity and explicit anaphoric references (e.g., maintaining previously established poses or constraints). Details are provided in Appendix~\ref{app:annotation}.

\subsubsection{Anaphora Detection}
\label{sec:anaphora_detection}
We define a segment as anaphoric if its motion constraints become more specific when conditioned on preceding segments (e.g., a body-part constraint, held object state, or pose that must persist across segments). Based on this definition, we used GPT-4o~\citep{gpt-4o} to determine whether each segment in the constructed sequences has a dependency and to identify its dependency source segments. For each target segment, the annotations of preceding segments were provided, and we prompted GPT-4o to perform binary anaphora detection to determine whether the motion described in the target segment becomes more specific when the preceding context is available, while also predicting the corresponding dependency source segment. This procedure was repeated three times, and a segment was regarded as anaphoric only when both the binary detection result and the identified dependency source segments were consistent across all three trials.

\subsection{Data splits and dataset contents}
\label{sec:data_splits_and_release}
To support reproducible benchmarking, we provide an official split of ELMA into 10.2k training clips and 3k test clips.
Each clip includes (i) a 30-second 3D motion sequence, (ii) six 5-second segment captions that consider preceding context, and (iii) segment-level dependency metadata indicating anaphoric target segments and their dependency source segments.
These components enable training and evaluation under both context-independent and context-dependent (anaphoric) multi-segment instruction formats.

\section{ELMABench}
\label{sec:bench}
\subsection{Benchmark overview}
ELMABench evaluates long-term text-to-motion generation under anaphoric multi-segment instructions, where later segments depend on earlier context.
ELMABench is constructed by randomly sampling 1k clips from the test split of the ELMA dataset (Sec.~\ref{sec:data_splits_and_release}).
Each benchmark instance provides (i) a sequence of six segment-level instructions with fixed durations and (ii) ground-truth motion for the full 30-second clip.
Crucially, evaluation is performed only on segments annotated as anaphoric targets, so that scores directly reflect a model's ability to preserve and propagate previously established constraints.

\subsection{Task setting}
Each test case in ELMABench corresponds to one 30-second motion clip, uniformly partitioned into six 5-second segments.
The input is a sequence of segment instructions $\mathbf{x}_{1:6}$ together with segment lengths $\boldsymbol{\ell}_{1:6}$, and the model outputs a single continuous motion sequence $\hat{\mathbf{m}}_{1:N}$.

We adopt the 263D motion representation of HumanML3D~\citep{humanml3d} for compatibility with standard evaluators.
Let $\mathcal{T}$ denote the set of target segment indices that are labeled as anaphoric (Sec.~\ref{sec:anaphora_detection}).
All metrics are computed by slicing the generated and ground-truth motions to the target segments $\{(\hat{\mathbf{m}}^{(t)}, \mathbf{m}^{(t)})\}_{t\in\mathcal{T}}$ and aggregating scores over $t\in\mathcal{T}$.
This prevents performance on non-anaphoric (easier) segments from diluting the benchmark signal.

\subsection{Evaluation settings: independent vs.\ anaphoric instructions}
To systematically analyze how models handle cross-segment dependencies, we consider two instruction formats:

\noindent \textbf{T2M (independent Text-to-Motion):} The standard setting where text instructions are strictly independent. Each segment description is self-contained without referencing prior states.

\noindent \textbf{A2M (Anaphoric Text-to-Motion):} Our proposed continuous setting where instructions exhibit anaphoric compositions. The model must resolve cross-dependencies and maintain constraints established in preceding segments.

Using these formats, we report three configurations to isolate the effect of ELMA supervision:
\textbf{(1) T2M Train $\rightarrow$ T2M Inference} serves as the standard baseline, evaluating the model's original generative capacity without complex dependencies, 
\textbf{(2) T2M Train $\rightarrow$ A2M Inference} evaluates the zero-shot generalization of standard models when subjected to anaphoric instructions, exposing their inherent limitations in state propagation, 
and \textbf{(3) A2M Train $\rightarrow$ A2M Inference} evaluates models after fine-tuning on our ELMA training set, demonstrating their acquired capacity to explicitly learn and preserve anaphoric constraints.

\subsection{Metrics}
To evaluate motion quality, we report four widely used metrics. FID measures the Fréchet distance between the distribution of generated motions and that of the ground-truth (GT) motions. R-Precision@3 indicates how highly the generated motions are ranked with respect to the corresponding GT descriptions through retrieval within a mini-batch. In addition, Matching Score computes the distance between text and motion embedding vectors to evaluate their alignment. Finally, Diversity measures the variance of the generated motion features across all sequences, reflecting the model's ability to produce varied movements.

\section{Experiments}
\begin{table*}[t]
\centering
\vspace{-0.6em}
\caption{
\textbf{Comparison under different training and inference settings.}
$\uparrow$ indicates higher is better, $\downarrow$ indicates lower is better, and $\rightarrow$ indicates closer to GT is better.
}
\vspace{-0.6em}
\label{tab:main_results}
\footnotesize
\renewcommand{\arraystretch}{1.08}
\setlength{\tabcolsep}{4.5pt}
\begin{tabular}{@{}llcccc@{}}
\toprule
Method & Setting & FID $\downarrow$ & R-P@3 $\uparrow$ & Diversity $\rightarrow$ & Matching Score $\downarrow$ \\
\midrule
GT
& --
& 0.21$_{\pm 0.03}$
& 0.480$_{\pm 0.009}$
& 7.31$_{\pm 0.16}$
& 4.53$_{\pm 0.07}$ \\

\midrule
\multirow{3}{*}{DoubleTake~\citep{doubletake}}
& T2M $\rightarrow$ T2M
& 4.64$_{\pm 0.20}$
& 0.189$_{\pm 0.007}$
& 8.07$_{\pm 0.09}$
& 12.38$_{\pm 0.02}$ \\
& T2M $\rightarrow$ A2M
& 5.16$_{\pm 0.23}$
& 0.184$_{\pm 0.006}$
& \textbf{7.91}$_{\pm 0.12}$
& \textbf{12.18}$_{\pm 0.03}$ \\
& A2M $\rightarrow$ A2M
& \textbf{3.95}$_{\pm 0.18}$
& \textbf{0.192}$_{\pm 0.007}$
& 8.10$_{\pm 0.11}$
& 12.34$_{\pm 0.04}$ \\

\midrule
\multirow{3}{*}{FlowMDM~\citep{flowmdm}}
& T2M $\rightarrow$ T2M
& 1.63$_{\pm 0.08}$
& 0.441$_{\pm 0.008}$
& 7.72$_{\pm 0.09}$
& 5.21$_{\pm 0.03}$ \\
& T2M $\rightarrow$ A2M
& 1.94$_{\pm 0.09}$
& 0.421$_{\pm 0.006}$
& \textbf{7.60}$_{\pm 0.10}$
& 5.32$_{\pm 0.03}$ \\
& A2M $\rightarrow$ A2M
& \textbf{1.53}$_{\pm 0.06}$
& \textbf{0.466}$_{\pm 0.008}$
& 8.12$_{\pm 0.11}$
& \textbf{4.92}$_{\pm 0.03}$ \\

\midrule
\multirow{3}{*}{UniMotion~\citep{unimotion}}
& T2M $\rightarrow$ T2M
& \textbf{18.24}$_{\pm 0.20}$
& \textbf{0.197}$_{\pm 0.004}$
& 10.10$_{\pm 0.02}$
& 6.47$_{\pm 0.06}$ \\
& T2M $\rightarrow$ A2M
& 18.35$_{\pm 0.20}$
& 0.192$_{\pm 0.005}$
& \textbf{10.03}$_{\pm 0.03}$
& \textbf{6.45}$_{\pm 0.08}$ \\
& A2M $\rightarrow$ A2M
& 19.01$_{\pm 0.15}$
& 0.185$_{\pm 0.003}$
& 10.11$_{\pm 0.02}$
& 6.52$_{\pm 0.06}$ \\
\bottomrule
\end{tabular}
\end{table*}


\begin{table*}[t]
\centering
\vspace{-0.1em}
\caption{
\textbf{Comparison between zero-shot inference and T2M-trained models.}
$\uparrow$ indicates higher is better, $\downarrow$ indicates lower is better, and $\rightarrow$ indicates closer to GT is better.
}
\vspace{-0.6em}
\label{tab:t2m_zeroshot}
\footnotesize
\renewcommand{\arraystretch}{1.08}
\setlength{\tabcolsep}{4.5pt}
\begin{tabular}{@{}llcccc@{}}
\toprule
Method & Setting & FID $\downarrow$ & R-P@3 $\uparrow$ & Diversity $\rightarrow$ & Matching Score $\downarrow$ \\
\midrule
GT
& --
& 0.21$_{\pm 0.03}$
& 0.480$_{\pm 0.009}$
& 7.31$_{\pm 0.16}$
& 4.53$_{\pm 0.07}$ \\

\midrule
\multirow{2}{*}{DoubleTake~\citep{doubletake}}
& Zero-shot
& 11.34$_{\pm 0.20}$
& 0.124$_{\pm 0.004}$
& 8.85$_{\pm 0.10}$
& 13.54$_{\pm 0.04}$ \\
& T2M
& \textbf{4.64}$_{\pm 0.20}$
& \textbf{0.189}$_{\pm 0.007}$
& \textbf{8.07}$_{\pm 0.09}$
& \textbf{12.38}$_{\pm 0.02}$ \\

\midrule
\multirow{2}{*}{FlowMDM~\citep{flowmdm}}
& Zero-shot
& 6.60$_{\pm 0.15}$
& 0.234$_{\pm 0.007}$
& 6.84$_{\pm 0.08}$
& 6.62$_{\pm 0.03}$ \\
& T2M
& \textbf{1.63}$_{\pm 0.08}$
& \textbf{0.441}$_{\pm 0.008}$
& \textbf{7.72}$_{\pm 0.09}$
& \textbf{5.21}$_{\pm 0.03}$ \\

\midrule
\multirow{2}{*}{UniMotion~\citep{unimotion}}
& Zero-shot
& 21.69$_{\pm 0.42}$
& 0.105$_{\pm 0.003}$
& 11.38$_{\pm 0.03}$
& \textbf{3.76}$_{\pm 0.06}$ \\
& T2M
& \textbf{18.24}$_{\pm 0.20}$
& \textbf{0.197}$_{\pm 0.004}$
& \textbf{10.10}$_{\pm 0.02}$
& 6.47$_{\pm 0.06}$ \\

\bottomrule
\end{tabular}
\vspace{-1.0em}
\end{table*}

\subsection{Baseline Methods}
\label{sec:Baseline}
We compare three representative models~\citep{doubletake, flowmdm, unimotion} in long-term motion generation as our baselines. To adapt them to our specific task setting, we apply minimal necessary modifications. For a fair comparison, all models utilize CLIP as their text encoder. Under the Anaphoric Text-to-Motion (A2M) setting, the text condition for each baseline is modified to take the weighted sum of the current segment's text embedding and all preceding segments' text embeddings. This modification explicitly injects historical context into the current generation step, enabling the models to maintain awareness of previously established constraints without requiring complex architectural redesigns. All baselines are carefully retrained to accommodate this contextual input setting.

\subsection{Effectiveness of ELMA for Anaphoric Text-to-Motion}
\label{sec:anaphoric_text_to_motion}
Tab.~\ref{tab:main_results} evaluates whether ELMA provides effective supervision for resolving cross-segment dependencies.
For each baseline, we compare three settings: standard T2M training and inference, T2M training with A2M inference, and A2M training with A2M inference.

For DoubleTake~\citep{doubletake} and FlowMDM~\citep{flowmdm}, A2M training with A2M inference achieves the best FID and R-Precision@3 within each model family, outperforming the corresponding T2M setting.
This suggests that ELMA contains useful supervision for learning cross-segment dependencies, and that models can exploit this information when trained and evaluated with the A2M format.
In contrast, applying A2M inference to T2M-trained models degrades performance, indicating that simply providing additional context at inference time is insufficient.

UniMotion is an exception to this trend. One possible explanation is that the principal component analysis (PCA) used to compress its frame-level text embeddings may discard information important for resolving cross-segment dependencies.
To examine this possibility, we conduct an ablation study varying the PCA dimensionality.
However, increasing the dimensionality does not improve performance, suggesting that PCA compression alone is unlikely to explain the degradation.
Another possible explanation is that UniMotion's frame-level text prediction loss becomes more challenging to optimize when the text embeddings encode richer contextual information.
We leave further investigation of this behavior to future work.

Overall, ELMA provides effective training signals for Anaphoric Text-to-Motion, but the benefit also depends on whether the model architecture can preserve and utilize long-range textual context.

\subsection{Effectiveness of ELMA for In-Domain Text-to-Motion}
Tab.~\ref{tab:t2m_zeroshot} evaluates whether ELMA also improves in-domain motion generation under the standard Text-to-Motion setting.
For each baseline, we compare a zero-shot model pretrained on HumanML3D~\citep{humanml3d} with the same model further trained on the ELMA training set using standard T2M inputs.

Across all baselines, ELMA training substantially improves FID and R-Precision@3 over the zero-shot setting.
This shows that the ELMA training set is not only useful for learning anaphoric dependencies, but also provides effective in-domain supervision for text-conditioned motion generation.
Although Matching Score does not improve uniformly, especially for FlowMDM, we treat it as an auxiliary metric in this benchmark.
Since many evaluation samples are originally constructed as multi-segment episodes, isolated segment-level text matching may not fully capture whether the generated motion satisfies the intended context-dependent behavior.
\subsection{Qualitative Results on Anaphoric Text-to-Motion}
\begin{figure*}[t]
    \centering

    \begin{subfigure}[t]{0.48\textwidth}
        \centering
        \includegraphics[width=\linewidth]{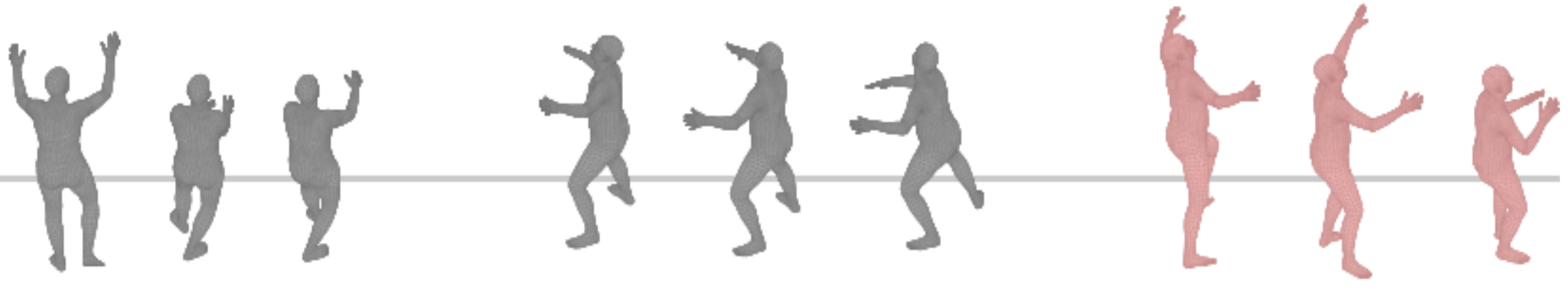}
        \caption{Ground truth}
        \label{fig:flowmdm_gt}
    \end{subfigure}
    \hfill
    \begin{subfigure}[t]{0.48\textwidth}
        \centering
        \includegraphics[width=\linewidth]{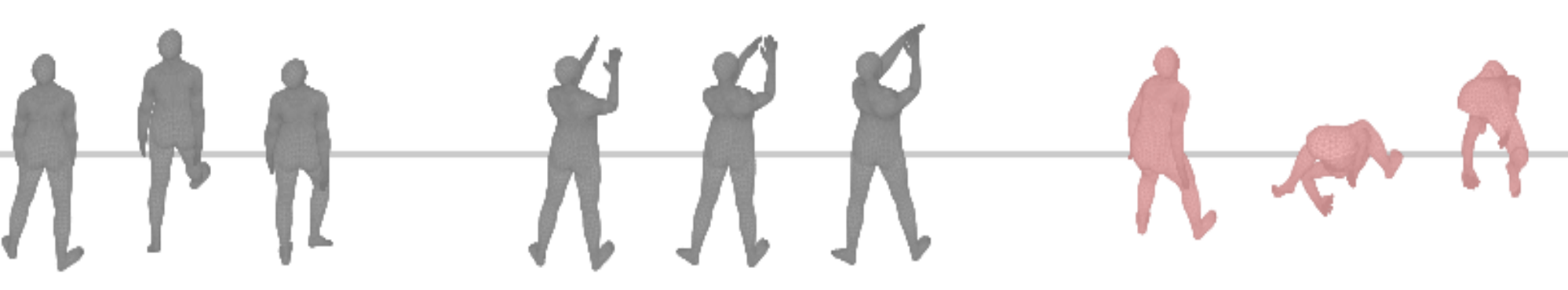}
        \caption{FlowMDM: Zero-shot}
        \label{fig:flowmdm_zeroshot}
    \end{subfigure}

    \vspace{0.25em}

    \begin{subfigure}[t]{0.48\textwidth}
        \centering
        \includegraphics[width=\linewidth]{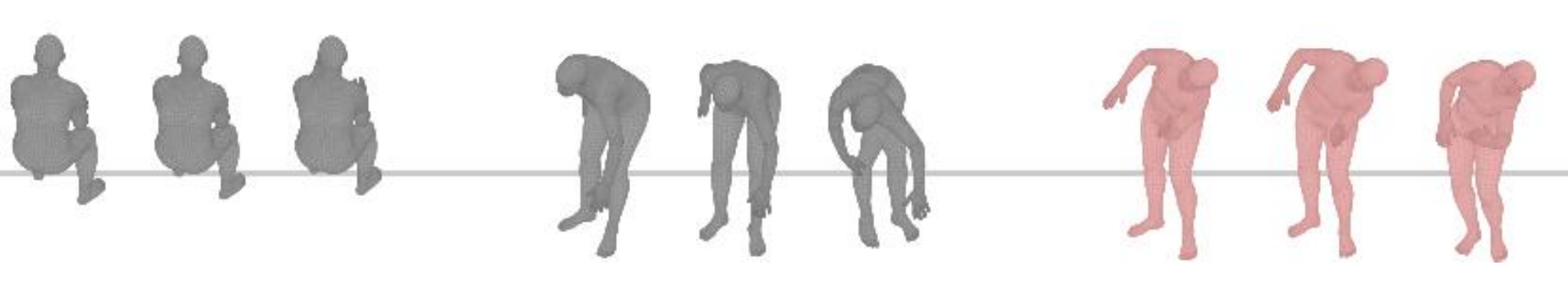}
        \caption{FlowMDM: T2M $\rightarrow$ T2M}
        \label{fig:flowmdm_t2m_t2m}
    \end{subfigure}
    \hfill
    \begin{subfigure}[t]{0.48\textwidth}
        \centering
        \includegraphics[width=\linewidth]{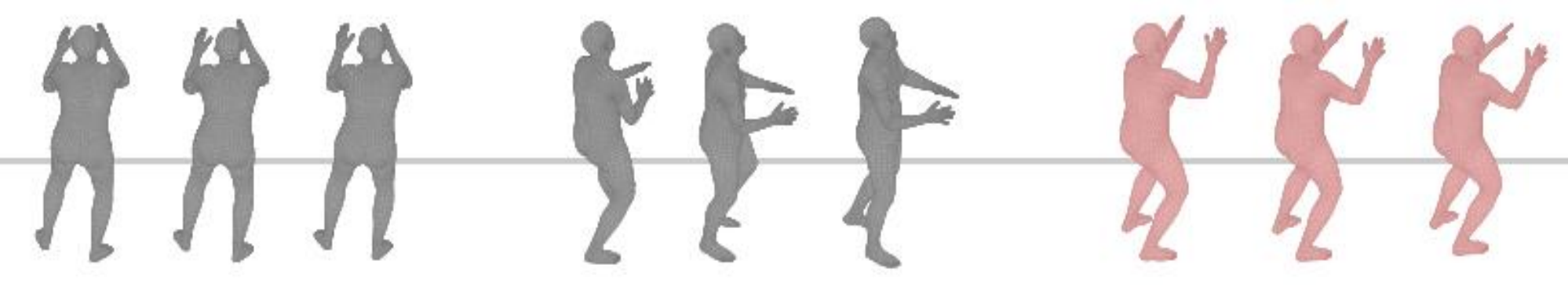}
        \caption{FlowMDM: A2M $\rightarrow$ A2M}
        \label{fig:flowmdm_a2m_a2m}
    \end{subfigure}

    \vspace{0.35em}

    \begin{subfigure}[t]{0.82\textwidth}
        \centering
        \includegraphics[width=\linewidth]{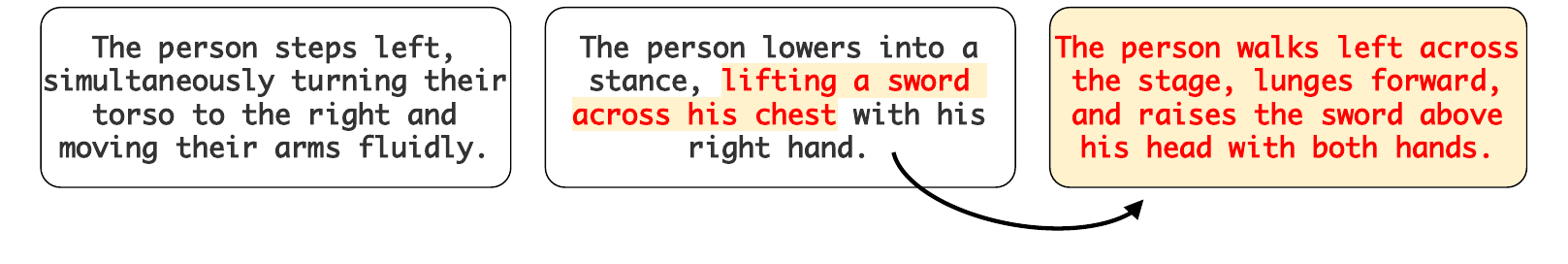}
        \caption{Input multi-segment instructions}
        \label{fig:flowmdm_instruction}
    \end{subfigure}

    \vspace{-0.25em}

    \caption{
    \textbf{Qualitative comparison of FlowMDM under different training and inference settings.}
    The bottom panel shows the input multi-segment instructions, where highlighted text indicates anaphoric dependencies.
    }
    \vspace{-0.9em}
    \label{fig:flowmdm_qualitative}
\end{figure*}
Fig.~\ref{fig:flowmdm_qualitative} shows qualitative results on a test case.
In the input shown in Fig.~\ref{fig:flowmdm_instruction}, the phrase ``lifting a sword across his chest'' in the preceding segment provides the context needed to interpret the following segment as lifting the sword from the chest.
The zero-shot and T2M training/T2M inference settings fail to resolve this cross-segment dependency.
In contrast, FlowMDM under the A2M training/A2M inference setting uses the context established in the preceding segment and generates motion that better follows the instruction.

\section{Conclusion and Limitations}
\label{sec:Limitations}
This paper introduced anaphoric text-to-motion, a long-term motion generation setting in which later instructions depend on entities, states, or constraints established in earlier segments. To support this setting, we presented the ELMA dataset and ELMABench for evaluating whether models can resolve anaphoric dependencies and preserve dependency-induced states over time.

While our framework demonstrates the potential of anaphoric long-term motion generation, several challenges remain. One limitation lies in the accuracy of VLM-based automatic annotation. Although recent VLMs can describe human motions in videos with increasing detail, they can still produce errors in fine-grained motion descriptions, such as confusing left and right hands or missing subtle object interactions. However, we believe these limitations will be alleviated by future advances in video understanding models, leading to more reliable and semantically consistent annotations.
Another important future direction is the development of scalable and cost-effective annotation pipelines for cross-segment dependencies. Efficiently generating high-quality annotations that naturally preserve inter-segment dependencies would further expand ELMA and facilitate larger-scale studies on context-aware motion generation.

Despite these limitations, we believe this work is an important first step toward addressing the challenging problem of maintaining temporal consistency in long-term text-to-motion generation. We hope ELMA and ELMABench will serve as foundations for future research in this direction.

\begin{ack}
We would like to thank Yuma Ono, Yugo Nishiyama, and Yu Hirose for their valuable contributions to data annotation and engineering support during the development of our dataset.

This work was partially supported by JSPS Grant-in-Aid for Challenging Research (Exploratory) 24K22296 and JST FOREST Program JPMJFR242I. Dixin Yang was supported by JST SPRING (Grant Number JPMJSP2123), and Kotaro Amaya was supported by JST BOOST (Grant Number JPMJBS2409).

\end{ack}
\bibliographystyle{plain}
\bibliography{main}

@String(PAMI  = {IEEE Trans. Pattern Anal. Mach. Intell.})

@String(CVPR  = {IEEE Conf. Comput. Vis. Pattern Recog.})

@String(ICCV  = {Int. Conf. Comput. Vis.})

@String(ECCV  = {Eur. Conf. Comput. Vis.})

@String(NeurIPS = {Adv. Neural Inform. Process. Syst.})

@String(ICML  = {Int. Conf. Mach. Learn.})

@String(ICLR  = {Int. Conf. Learn. Represent.})

@String(CVPRW = {IEEE Conf. Comput. Vis. Pattern Recog. Worksh.})

@String(PAMI  = {IEEE TPAMI})

@String(CVPR  = {CVPR})

@String(ICCV  = {ICCV})

@String(ECCV  = {ECCV})

@String(NeurIPS = {NeurIPS})

@String(ICML  = {ICML})

@String(ICLR  = {ICLR})

@String(CVPRW = {CVPRW})

@inproceedings{mdm,
  title={Human Motion Diffusion Model},
  author={Tevet, Guy and Raab, Sigal and Gordon, Brian and Shafir, Yoni and Cohen-Or, Daniel and Bermano, Amit Haim},
  booktitle=ICLR,
  year={2023}
}

@inproceedings{t2m-gpt,
  title={Generating human motion from textual descriptions with discrete representations},
  author={Zhang, Jianrong and Zhang, Yangsong and Cun, Xiaodong and Zhang, Yong and Zhao, Hongwei and Lu, Hongtao and Shen, Xi and Shan, Ying},
  booktitle=CVPR,
  pages={14730--14740},
  year={2023}
}

@article{motiondiffuse,
  title={Motiondiffuse: Text-driven human motion generation with diffusion model},
  author={Zhang, Mingyuan and Cai, Zhongang and Pan, Liang and Hong, Fangzhou and Guo, Xinying and Yang, Lei and Liu, Ziwei},
  journal=PAMI,
  volume={46},
  number={6},
  pages={4115--4128},
  year={2024},
}

@inproceedings{momask,
  title={Momask: Generative masked modeling of 3d human motions},
  author={Guo, Chuan and Mu, Yuxuan and Javed, Muhammad Gohar and Wang, Sen and Cheng, Li},
  booktitle=CVPR,
  pages={1900--1910},
  year={2024}
}

@inproceedings{motionctrl,
  title={MotionCtrl: A Real-time Controllable Vision-Language-Motion Model},
  author={Cao, Bin and Zheng, Sipeng and Wang, Ye and Xia, Lujie and Wei, Qianshan and Jin, Qin and Liu, Jing and Lu, Zongqing},
  booktitle=ICCV,
  pages={12253--12262},
  year={2025}
}

@inproceedings{motionstreamer,
  title={Motionstreamer: Streaming motion generation via diffusion-based autoregressive model in causal latent space},
  author={Xiao, Lixing and Lu, Shunlin and Pi, Huaijin and Fan, Ke and Pan, Liang and Zhou, Yueer and Feng, Ziyong and Zhou, Xiaowei and Peng, Sida and Wang, Jingbo},
  booktitle=ICCV,
  pages={10086--10096},
  year={2025}
}

@inproceedings{doubletake,
  title={Human Motion Diffusion as a Generative Prior},
  author={Shafir, Yoni and Tevet, Guy and Kapon, Roy and Bermano, Amit Haim},
  booktitle=ICLR,
  year={2024}
}

@inproceedings{flowmdm,
  title={Seamless human motion composition with blended positional encodings},
  author={Barquero, German and Escalera, Sergio and Palmero, Cristina},
  booktitle=CVPR,
  pages={457--469},
  year={2024}
}

@inproceedings{dartcontrol,
  title={DartControl: A Diffusion-Based Autoregressive Motion Model for Real-Time Text-Driven Motion Control},
  author={Zhao, Kaifeng and Li, Gen and Tang, Siyu},
  booktitle=ICLR,
  year = {2025}
}

@inproceedings{t2lm,
  title={T2lm: Long-term 3d human motion generation from multiple sentences},
  author={Lee, Taeryung and Baradel, Fabien and Lucas, Thomas and Lee, Kyoung Mu and Rogez, Gr{\'e}gory},
  booktitle=CVPRW,
  pages={1867--1876},
  year={2024}
}

@inproceedings{stmc,
  title={Multi-track timeline control for text-driven 3d human motion generation},
  author={Petrovich, Mathis and Litany, Or and Iqbal, Umar and Black, Michael J and Varol, Gul and Bin Peng, Xue and Rempe, Davis},
  booktitle=CVPRW,
  pages={1911--1921},
  year={2024}
}

@inproceedings{transphase,
  title={Deep Compositional Phase Diffusion for Long Motion Sequence Generation},
  author={Au, Ho Yin and Chen, Jie and Jiang, Junkun and Xiang, Jingyu},
  booktitle=NeurIPS,
  year={2025}
}

@inproceedings{unimotion,
  title={Unimotion: Unifying 3d human motion synthesis and understanding},
  author={Li, Chuqiao and Chibane, Julian and He, Yannan and Pearl, Naama and Geiger, Andreas and Pons-Moll, Gerard},
  booktitle={3DV},
  pages={240--249},
  year={2025},
}

@inproceedings{humanml3d,
  title={Generating diverse and natural 3d human motions from text},
  author={Guo, Chuan and Zou, Shihao and Zuo, Xinxin and Wang, Sen and Ji, Wei and Li, Xingyu and Cheng, Li},
  booktitle=CVPR,
  pages={5152--5161},
  year={2022}
}

@inproceedings{amass,
  title={AMASS: Archive of motion capture as surface shapes},
  author={Mahmood, Naureen and Ghorbani, Nima and Troje, Nikolaus F and Pons-Moll, Gerard and Black, Michael J},
  booktitle=ICCV,
  pages={5442--5451},
  year={2019}
}

@article{motion-x,
  title={Motion-x: A large-scale 3d expressive whole-body human motion dataset},
  author={Lin, Jing and Zeng, Ailing and Lu, Shunlin and Cai, Yuanhao and Zhang, Ruimao and Wang, Haoqian and Zhang, Lei},
  journal=NeurIPS,
  volume={36},
  pages={25268--25280},
  year={2023}
}

@inproceedings{babel,
  title={BABEL: Bodies, action and behavior with english labels},
  author={Punnakkal, Abhinanda R and Chandrasekaran, Arjun and Athanasiou, Nikos and Quiros-Ramirez, Alejandra and Black, Michael J},
  booktitle=CVPR,
  pages={722--731},
  year={2021}
}

@inproceedings{teach,
  title={Teach: Temporal action composition for 3d humans},
  author={Athanasiou, Nikos and Petrovich, Mathis and Black, Michael J and Varol, G{\"u}l},
  booktitle={3DV},
  pages={414--423},
  year={2022},
}

@article{finemogen,
  title={Finemogen: Fine-grained spatio-temporal motion generation and editing},
  author={Zhang, Mingyuan and Li, Huirong and Cai, Zhongang and Ren, Jiawei and Yang, Lei and Liu, Ziwei},
  journal=NeurIPS,
  volume={36},
  pages={13981--13992},
  year={2023}
}

@inproceedings{dmc,
  title={Dense Motion Captioning},
  author={Xu, Shiyao and Liberatori, Benedetta and Varol, G{\"u}l and Rota, Paolo},
  booktitle={2026 International Conference on 3D Vision (3DV)},
  pages={1750--1759},
  year={2026},
}

@article{frankenmotion,
  title={FrankenMotion: Part-level Human Motion Generation and Composition},
  author={Li, Chuqiao and Xie, Xianghui and Cao, Yong and Geiger, Andreas and Pons-Moll, Gerard},
  journal={arXiv preprint arXiv:2601.10909},
  year={2026}
}

@inproceedings{finemotion,
  title={FineMotion: A Dataset and Benchmark with both Spatial and Temporal Annotation for Fine-grained Motion Generation and Editing},
  author={Wu, Bizhu and Xie, Jinheng and Ding, Meidan and Kong, Zhe and Ren, Jianfeng and Bai, Ruibin and Qu, Rong and Shen, Linlin},
  booktitle=ICCV,
  pages={13837--13846},
  year={2025}
}

@inproceedings{yolo-pose,
  title={Yolo-pose: Enhancing yolo for multi person pose estimation using object keypoint similarity loss},
  author={Maji, Debapriya and Nagori, Soyeb and Mathew, Manu and Poddar, Deepak},
  booktitle=CVPRW,
  pages={2637--2646},
  year={2022}
}

@article{posescript,
  title={Posescript: Linking 3d human poses and natural language},
  author={Delmas, Ginger and Weinzaepfel, Philippe and Lucas, Thomas and Moreno-Noguer, Francesc and Rogez, Gr{\'e}gory},
  journal=PAMI,
  year={2025},
  pages={5146--5159},
}

@inproceedings{bytetrack,
  title={Bytetrack: Multi-object tracking by associating every detection box},
  author={Zhang, Yifu and Sun, Peize and Jiang, Yi and Yu, Dongdong and Weng, Fucheng and Yuan, Zehuan and Luo, Ping and Liu, Wenyu and Wang, Xinggang},
  booktitle=ECCV,
  pages={1--21},
  year={2022},
}

@inproceedings{sam2,
  title={{SAM 2}: Segment Anything in Images and Videos},
  author={Ravi, Nikhila and Gabeur, Valentin and Hu, Yuan-Ting and Hu, Ronghang and Ryali, Chaitanya and Ma, Tengyu and Khedr, Haitham and R{\"a}dle, Roman and Rolland, Chloe and Gustafson, Laura and others},
  booktitle=ICLR,
  year={2025}
}

@inproceedings{gvhmr,
  title={World-grounded human motion recovery via gravity-view coordinates},
  author={Shen, Zehong and Pi, Huaijin and Xia, Yan and Cen, Zhi and Peng, Sida and Hu, Zechen and Bao, Hujun and Hu, Ruizhen and Zhou, Xiaowei},
  booktitle={SIGGRAPH Asia 2024},
  pages={1--11},
  year={2024}
}

@misc{gemini25pro,
  author       = {{Google DeepMind}},
  title        = {Gemini 2.5: Our most intelligent AI model},
  year         = {2025},
  howpublished = {\url{https://blog.google/innovation-and-ai/models-and-research/google-deepmind/gemini-model-thinking-updates-march-2025/}},
  note         = {Accessed: 2026-04-29}
}

@article{gpt-4o,
  title={GPT-4o System Card},
  author={{OpenAI}},
  journal={arXiv preprint arXiv:2410.21276},
  year={2024}
}

@inproceedings{opent2m,
  title={OpenT2M: No-frill Motion Generation with Open-source, Large-scale, High-quality Data},
  author={Cao, Bin and Zheng, Sipeng and Luo, Hao and Li, Boyuan and Liu, Jing and Lu, Zongqing},
  booktitle=CVPR,
  pages={30640--30649},
  year={2026}
}

@article{kinetics700,
  title={A Short Note on the Kinetics-700-2020 Human Action Dataset},
  author={Smaira, Lucas and Carreira, Jo{\~a}o and Noland, Eric and Clancy, Ellen and Wu, Amy and Zisserman, Andrew},
  journal={arXiv preprint arXiv:2010.10864},
  year={2020}
}

@article{qwen25vl,
  title={Qwen2.5-VL Technical Report},
  author={Bai, Shuai and Chen, Keqin and Liu, Xuejing and Wang, Jialin and Ge, Wenbin and Song, Sibo and Dang, Kai and Wang, Peng and Wang, Shijie and Tang, Jun and Zhong, Humen and Zhu, Yuanzhi and Yang, Mingkun and Li, Zhaohai and Wan, Jianqiang and Wang, Pengfei and Ding, Wei and Fu, Zheren and Xu, Yiheng and Ye, Jiabo and Zhang, Xi and Xie, Tianbao and Cheng, Zesen and Zhang, Hang and Yang, Zhibo and Xu, Haiyang and Lin, Junyang},
  journal={arXiv preprint arXiv:2502.13923},
  year={2025}
}

@inproceedings{
  posefix,
  title={PoseFix: Correcting 3D Human Poses with Natural Language},
  author={Delmas, Ginger and Weinzaepfel, Philippe and Moreno-Noguer, Francesc and Rogez, Gregory},
  booktitle=ICCV,
  pages={14972--14982},
  year={2023}
}

@inproceedings{clip,
  title={Learning transferable visual models from natural language supervision},
  author={Radford, Alec and Kim, Jong Wook and Hallacy, Chris and Ramesh, Aditya and Goh, Gabriel and Agarwal, Sandhini and Sastry, Girish and Askell, Amanda and Mishkin, Pamela and Clark, Jack and others},
  booktitle=ICML,
  pages={8748--8763},
  year={2021},
}

@inproceedings{smplx,
  title={Expressive body capture: 3d hands, face, and body from a single image},
  author={Pavlakos, Georgios and Choutas, Vasileios and Ghorbani, Nima and Bolkart, Timo and Osman, Ahmed AA and Tzionas, Dimitrios and Black, Michael J},
  booktitle=CVPR,
  pages={10975--10985},
  year={2019}
}

@inproceedings{hummanmogen,
  title     = {Humman: Multi-modal 4d human dataset for versatile sensing and modeling},
  author    = {Cai, Zhongang and Ren, Daxuan and Zeng, Ailing and Lin, Zhengyu and Yu, Tao and Wang, Wenjia and Fan, Xiangyu and Gao, Yang and Yu, Yifan and Pan, Liang and Hong, Fangzhou and Zhang, Mingyuan and Loy, Chen Change and Yang, Lei and Liu, Ziwei},
  booktitle = ECCV,
  pages     = {557--577},
  year      = {2022}
}

\clearpage
\appendix
\raggedbottom
\begin{center}
    {\Large\bfseries Appendix}
\end{center}

\etocsetlocaltop.toc{part}

\begingroup
\hypersetup{linkcolor=black}
\setcounter{tocdepth}{3}
\etocsettocstyle{\section*{Contents}}{}
\localtableofcontents
\endgroup

\section{Dataset Construction Pipeline Details}
\subsection{Motion Extraction}
\subsubsection{Online Video Search}
\label{app:search_queries}

The search queries for video retrieval were constructed based on the action categories from Kinetics-700~\citep{kinetics700} and the raw labels from BABEL~\citep{babel}.
We translated each label into Japanese, Chinese, and English, and further expanded the queries by using GPT-4o~\citep{gpt-4o} to generate five synonymous expressions for each language, aiming to broaden the source-video distribution and capture a diverse set of human motions.

This process resulted in a total of 7,592 search queries, distributed approximately equally across English, Japanese, and Chinese ($\approx$33\% each).
The most frequent search query accounted for only 0.2\% of the collected samples, while the top 10 queries collectively accounted for 1.20\%.
These statistics indicate that the collected samples are not dominated by a small number of search queries.

\subsubsection{Scene Segmentation}
Raw videos often undergo frequent scene cuts, which can undermine the consistency of motion data. We leverage PySceneDetect\footnote{\url{https://github.com/Breakthrough/PySceneDetect}} to segment the footage, retaining only video segments without detected scene cuts for at least 30 seconds. We choose 30 seconds as a practical trade-off between semantic transitions and data availability.
\subsubsection{Hierarchical Multi-Person Detection and Tracking}
\label{app:preprocessing}
This stage addresses the technical challenges of scaling motion extraction to long-term sequences.
First, error accumulation poses a significant risk in long-term mesh estimation. Continuous mesh estimation over extended sequences is prone to drift, especially when the subject undergoes occlusion or exits the frame. To mitigate this, it is crucial to filter out low-quality input segments in which the tracked subject is partially occluded or not fully visible at any point in the sequence.
Second, such high-quality segments are relatively scarce in online videos. Consequently, conventional single-subject filtering yields low data density, making large-scale collection inefficient.
To address these challenges, we adopt a hierarchical strategy: a coarse keypoint-based tracker for high-recall candidate mining, followed by a fine pixel-level mask propagation for high-precision cropping.

\noindent\textbf{Coarse stage: Candidate Selection and Filtering with Keypoint-based tracking} \\
We first perform a computationally coarse screening to identify suitable video segments using YOLO-Pose~\citep{yolo-pose} and ByteTrack~\citep{bytetrack}. YOLO-Pose provides per-frame person detection together with 2D body keypoints, and ByteTrack associates these detections across frames to form consistent person tracks. To maximize data yield, we extract all eligible subjects within a frame rather than limiting detection to a single primary actor.
We then sample frames at regular intervals, run pose estimation on the sampled frames, and filter the resulting tracks based on two criteria: 

\begin{itemize}
    \item \textbf{Keypoint Reliability.}
    To ensure robust pose estimation, we require that a significant proportion of body joints predicted by YOLO-Pose exhibit high per-joint confidence scores.
    This step effectively filters out subjects with partial visibility or severe occlusion.

    \item \textbf{Relative Movement Score.}
    To exclude near-static subjects, we compute a relative movement score for each track within a scene.
    Let $\mathbf{c}_t$ denote the bounding-box center at frame $t$, and let $a_t$ denote the bounding-box area.
    We define the score as
    \[
    d_{\mathrm{rel}}
    =
    \frac{
        \sum_{t=2}^{T} \|\mathbf{c}_t-\mathbf{c}_{t-1}\|_2
        +
        \lambda_{\mathrm{scale}}
        \sum_{t=2}^{T}
        \left|
        \sqrt{a_t}-\sqrt{a_{t-1}}
        \right|
    }{
        \sqrt{W^2+H^2}
    },
    \]
    where $W$ and $H$ are the width and height of the original video frame, and $\lambda_{\mathrm{scale}}$ is the weight for bounding-box scale changes.
    The first term captures image-plane displacement, while the second term accounts for apparent scale changes, which often reflect depth-wise motion.
    We discard a track as low-motion if $d_{\mathrm{rel}} < \gamma_{\mathrm{move}}$.
    This score is computed once over the entire track within a scene.
\end{itemize}

\noindent\textbf{Fine stage: Pixel-level BBox Refinement} \\
The bounding box from the initial frame of the valid clip serves as the initialization prompt for the fine stage. To obtain precise spatiotemporal localization required for high-fidelity mesh reconstruction, we leverage SAM 2~\citep{sam2} to propagate a segmentation mask from the first frame across the entire sequence. This pixel-level refinement ensures that the subject is consistently tracked with high precision, providing stable input crops that are essential for minimizing drift during the subsequent mesh estimation process.
\subsubsection{MLLM-based Verification}
Following the automated tracking pipeline, we apply Gemini 2.5 Pro~\citep{gemini25pro} to verify the collected clips and filter out failure cases that are difficult to remove using geometric or keypoint-based criteria alone. These cases mainly include non-human subjects, such as animated characters or statues, mirror reflections, and subtle occlusions that compromise the physical validity of the extracted motion.
\begin{table}[t]
\centering
\caption{
Parameters used in Motion Extraction
}
\label{tab:collection_hyperparams}
\footnotesize
\renewcommand{\arraystretch}{1.1}
\setlength{\tabcolsep}{4pt}
\begin{tabular}{lll}
\toprule
Stage & Hyperparameter & Value \\
\midrule
Tracking & Model checkpoint & YOLOv8l-pose \\
Tracking & Tracker & ByteTrack \\
Tracking & Sampling interval & 4 frames \\
\midrule
Filtering & Detection confidence threshold & 0.5 \\
Filtering & Keypoint confidence threshold & 0.3 \\
Filtering & Minimum clip duration & 30.0 s \\
Filtering & Minimum relative movement & 0.7 \\
Filtering & Scale-change weight & 2.0 \\
Filtering & Required reliable joints & 13 \\
\bottomrule
\end{tabular}
\end{table}

\subsection{Postprocessing}
\label{app:postprocessing}
\subsubsection{PoseFiltering}
For each clip, we compute a pose naturalness score based on VPoser~\citep{smplx}.
Let $\mathbf{x}_t \in \mathbb{R}^{21\times 3}$ denote the body pose at frame $t$, represented in axis-angle form.
We feed $\mathbf{x}_t$ into a pretrained VPoser encoder and obtain the latent mean $\mathbf{z}_t = E(\mathbf{x}_t)$, where $E(\cdot)$ denotes the VPoser encoder.
The reconstructed pose is then obtained as $\hat{\mathbf{x}}_t = D(\mathbf{z}_t)$, where $D(\cdot)$ denotes the VPoser decoder.
Since VPoser is trained on natural human poses, poses that deviate from the learned pose manifold tend to yield larger reconstruction errors.

To measure the reconstruction error, we compare the input and reconstructed poses in rotation space.
Let $\mathbf{x}_{t,j} \in \mathbb{R}^{3}$ and $\hat{\mathbf{x}}_{t,j} \in \mathbb{R}^{3}$ denote the axis-angle representations of joint $j$ at frame $t$ for the input and reconstructed poses, respectively.
We convert them into rotation matrices as $R_{t,j}=\mathrm{Rot}(\mathbf{x}_{t,j})$ and $\hat{R}_{t,j}=\mathrm{Rot}(\hat{\mathbf{x}}_{t,j})$, where $\mathrm{Rot}(\cdot)$ denotes the axis-angle to rotation-matrix conversion.
The joint-wise reconstruction error is defined as the geodesic rotation error:
\[
e_{t,j}
=
\arccos\!\left(
\frac{
\mathrm{tr}\!\left(R_{t,j}\hat{R}_{t,j}^{\top}\right)-1
}{2}
\right).
\]
This measures the angular distance between the input and reconstructed rotations.
We then average the joint-wise errors within each frame to obtain the frame-level reconstruction error
$e_t = \frac{1}{J}\sum_{j=1}^{J} e_{t,j}$, where $J$ is the number of body joints.

Next, we aggregate the frame-level errors over the entire clip using the 99th percentile, defined as
$s = \mathrm{Percentile}_{99}(e_1,\dots,e_T)$, where $T$ denotes the number of frames in the clip.
Using the 99th percentile makes the score sensitive to clips that contain a small number of severely corrupted poses.
Finally, we reject a clip if $s > \gamma^{\mathrm{pose}}$, where $\gamma^{\mathrm{pose}}$ is a predefined threshold.
This allows us to remove clips containing implausible poses while complementing temporal filtering based on motion smoothness.
\subsubsection{MotionFiltering+}
For each clip, we compute two temporal irregularity measures: joint-wise rotation changes and jerk.
Let $R_{t,j}$ denote the rotation matrix of joint $j$ at frame $t$.
We define the frame-to-frame rotation change of joint $j$ as
$\Delta \theta_{t,j} = \mathrm{angle}(R_{t,j}R_{t-1,j}^{\top})$.
This quantity becomes large when a joint undergoes an unusually abrupt orientation change between adjacent frames.

To capture temporal irregularities in motion trajectories, we also compute jerk from the estimated 3D joint positions.
Let $\mathbf{p}_{t,j} \in \mathbb{R}^{3}$ denote the 3D position of joint $j$ at frame $t$.
We define the frame-level jerk as the mean magnitude of the third-order temporal difference over joints:
\vspace{2pt}
\[
\mathrm{jerk}_t
=
\frac{1}{J}
\sum_{j=1}^{J}
\left\|
\Delta^3 \mathbf{p}_{t,j}
\right\|_2 .
\]
\vspace{2pt}
Intuitively, jerk measures rapid changes in acceleration and is therefore sensitive to jittery or physically implausible motion.

Using AMASS as a clean motion reference, we precompute the 99th-percentile thresholds for each joint-wise rotation change and for jerk, denoted by $\tau^{\mathrm{rot}}_j$ and $\tau^{\mathrm{jerk}}$, respectively.
Based on these thresholds, we define the outlier indicators as
\vspace{2pt}
\[
m^{\mathrm{rot}}_{t,j}
=
\mathds{1}\!\left[\Delta \theta_{t,j} > \tau^{\mathrm{rot}}_j\right],
\qquad
m^{\mathrm{jerk}}_{t}
=
\mathds{1}\!\left[\mathrm{jerk}_{t} > \tau^{\mathrm{jerk}}\right].
\]
\vspace{2pt}
We then aggregate these indicators over the entire clip:
\[
r^{\mathrm{rot}}
=
\frac{1}{T}\sum_{t=1}^{T}\frac{1}{J}\sum_{j=1}^{J} m^{\mathrm{rot}}_{t,j},
\qquad
r^{\mathrm{jerk}}
=
\frac{1}{T}\sum_{t=1}^{T} m^{\mathrm{jerk}}_{t}.
\]
Here, $r^{\mathrm{rot}}$ measures the average fraction of joints whose frame-to-frame rotation change is abnormal, while $r^{\mathrm{jerk}}$ measures the frequency of temporally irregular frames over the clip.

Finally, a clip is rejected if
\vspace{2pt}
\[
r^{\mathrm{rot}} > \gamma_{\mathrm{rot}}
\quad \text{or} \quad
r^{\mathrm{jerk}} > \gamma_{\mathrm{jerk}},
\]
\vspace{3pt}
where $\gamma_{\mathrm{rot}}$ and $\gamma_{\mathrm{jerk}}$ are predefined clip-level thresholds.
In addition, we remove near-static failure cases where the reconstructed root translation is negligible over the entire clip, i.e., $d^{\mathrm{trans}} < \gamma_{\mathrm{trans}}$, which typically indicates tracking or reconstruction collapse.

\begin{table}[t]
\centering
\caption{
Parameters used in Postprocessing
}
\label{tab:postprocessing_hyperparams}
\footnotesize
\renewcommand{\arraystretch}{1.1}
\setlength{\tabcolsep}{4pt}
\begin{tabular}{lll}
\toprule
Stage & Hyperparameter & Value \\
\midrule
MotionFiltering+ & Rotation outlier ratio threshold $\gamma_{\mathrm{rot}}$ & 0.4 \\
MotionFiltering+ & Jerk outlier rate threshold $\gamma_{\mathrm{jerk}}$ & 0.5 \\
MotionFiltering+ & Low-motion translation threshold $\gamma_{\mathrm{trans}}$ & 2.0 \\
\bottomrule
\end{tabular}
\end{table}

\section{Dataset Details}
\label{app:dataset_details}
\subsection{Semantic Diversity}
\begin{figure}
    \centering
    \includegraphics[width=1.0\linewidth]{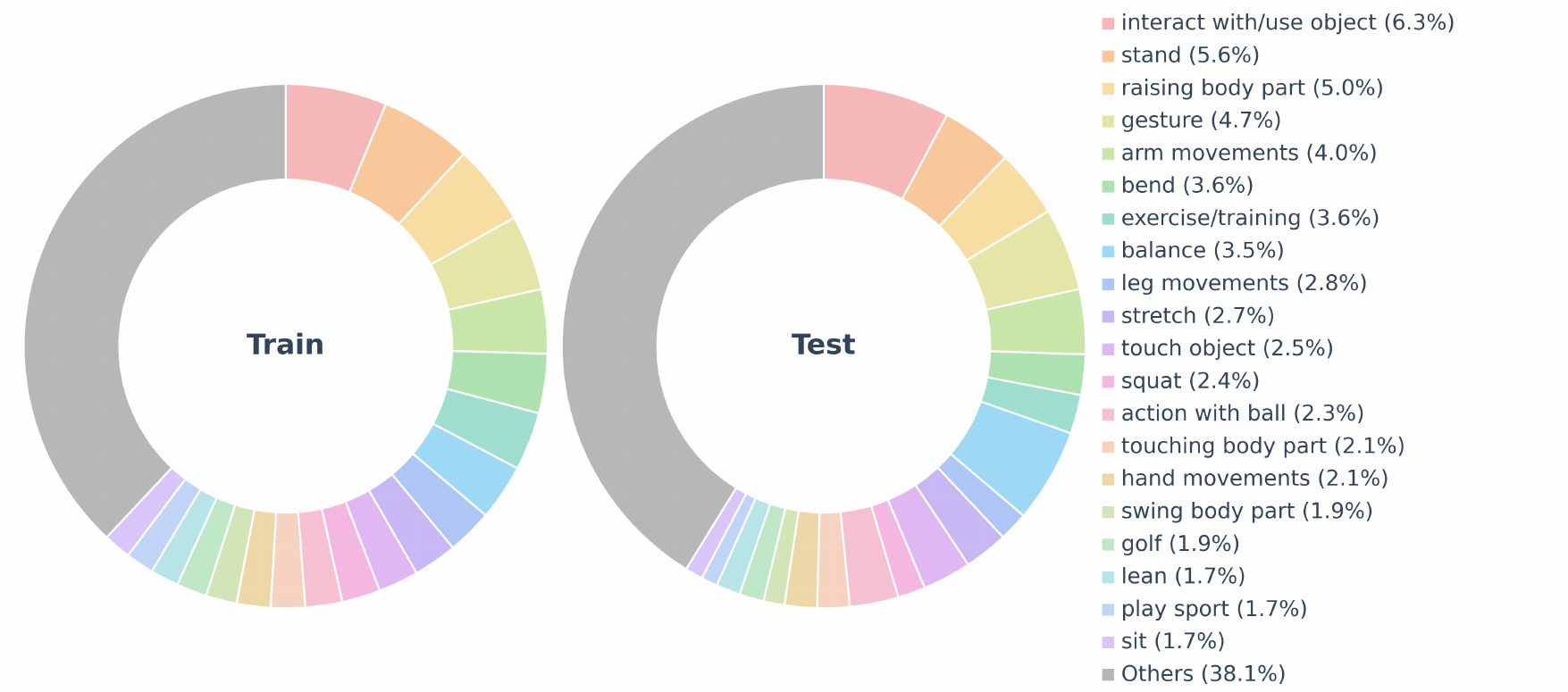}
    \caption{
    Analysis of the ELMA dataset. Each category corresponds to an action label from BABEL.
}
    \label{fig:semantic_diversity}
\end{figure}

To analyze the semantic coverage of the ELMA dataset, we illustrate the distribution of action categories in Fig.~\ref{fig:semantic_diversity}. The categories are mapped from BABEL action labels, covering a wide spectrum of human movements. As shown in the figure, both the training and test splits demonstrate a highly consistent and diverse distribution of activities. These range from object-centric tasks (e.g., interact with/use object) and fundamental postures (e.g., stand) to dynamic sports and fitness routines (e.g., golf). This broad distribution confirms that ELMA effectively captures the physical variability of real-world episodic activities.

\subsection{Postprocessing Analysis}
\begin{figure}
    \centering
    \begin{subfigure}[t]{0.48\linewidth}
        \centering
        \includegraphics[width=\linewidth]{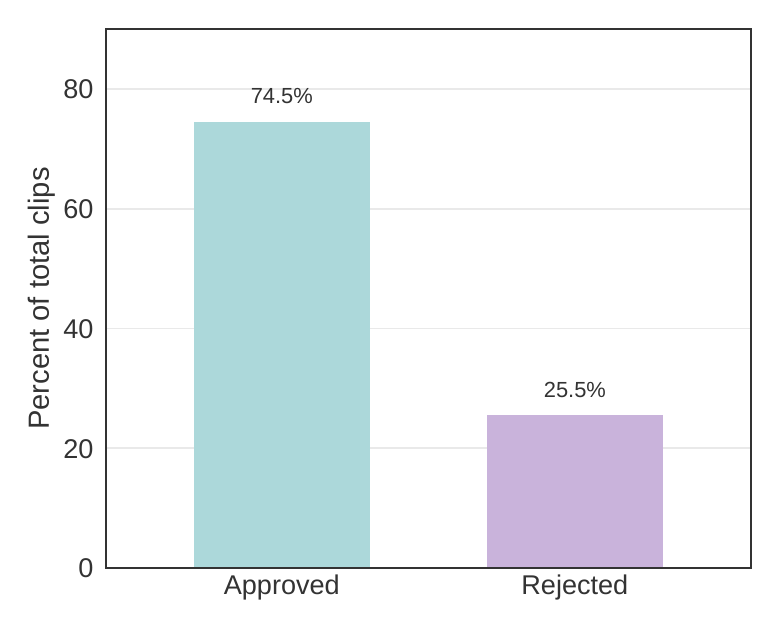}
        \caption{Overall acceptance rate}
        \label{fig:overall_accptance}
    \end{subfigure}
    \begin{subfigure}[t]{0.49\linewidth}
        \centering
        \includegraphics[width=\linewidth]{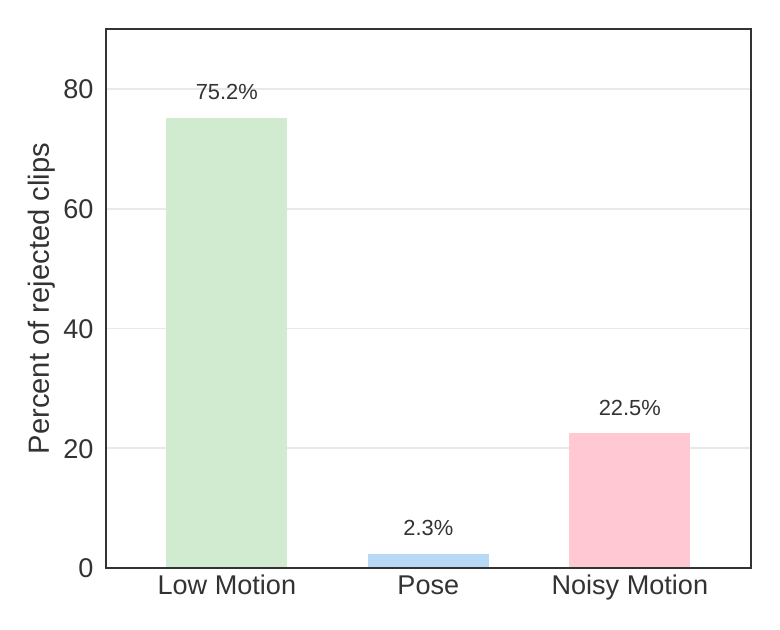}
        \caption{Filtering ratio at each postprocessing stage}
        \label{fig:filtering_ratio}
    \end{subfigure}
    \caption{
    Motion acceptance rate through postprocessing and the filtering ratio at each stage.
}
    \label{fig:postprocessing_appendix}
\end{figure}
\subsubsection{Filtering Ratios at Each Processing Stage}
Fig.~\ref{fig:postprocessing_appendix} shows the acceptance and rejection rates of motions before and after postprocessing, together with the breakdown of rejected motions. Approximately 25\% of all motions are rejected by postprocessing, and about 75\% of the rejected motions are identified as near-static failure cases. This suggests that, in in-the-wild videos, people who remain visible for a long duration within the same scene do not necessarily perform meaningful actions and are often static. Postprocessing detects and removes such cases at the motion level, which could not be fully filtered out during the preprocessing stage in Appendix~\ref{app:preprocessing}, thereby extracting samples that are more suitable as motion data.

\subsubsection{Effect of AMASS-based Filtering on Semantic Diversity}
PoseFiltering and MotionFiltering+ are both AMASS-based postprocessing steps.
Although these filters improve motion quality by removing implausible poses and temporally unstable motions, overly strong AMASS-based filtering could potentially bias the resulting dataset toward AMASS-like motion patterns and reduce semantic diversity.
To examine this possibility, we compare the vocabulary overlap between ELMA and HumanML3D after postprocessing. The vocabulary Jaccard similarity between ELMA and HumanML3D is 0.278, suggesting that the AMASS-based filtering steps remove noisy motions without collapsing ELMA into an AMASS-like text domain, thereby preserving semantic diversity with object- and scene-specific terms.

\subsection{Additional Dataset Samples}
\begin{figure}[H]
    \centering
    \includegraphics[width=\linewidth]{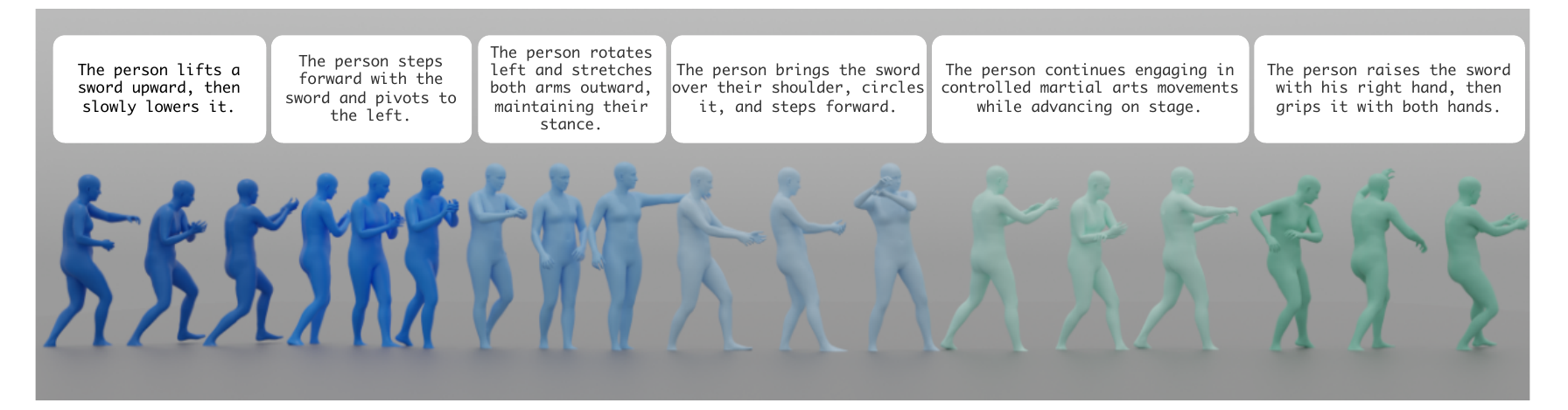}

    \vspace{0.5em}

    \includegraphics[width=\linewidth]{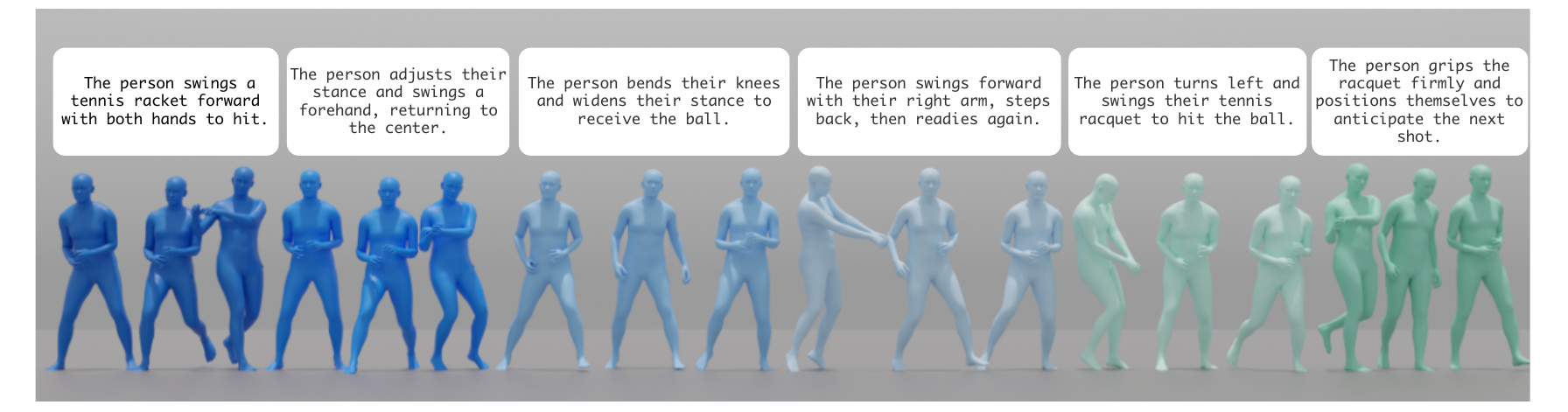}

    \caption{
        Additional examples from the ELMA dataset.
    }
    \label{fig:additional_samples}
\end{figure}

\section{Text Annotation and Anaphora Detection Details}
\label{app:annotation}
\subsection{Prompt Design}
\noindent
\noindent
\begin{minipage}[t]{0.48\textwidth}
\begin{tcblisting}{
  title=Prompt for Anaphora Detection,
  colback=gray!5,
  colframe=gray!60,
  fonttitle=\bfseries,
  listing only,
  height=0.78\textheight,
  valign=top,
  left=0.8mm,
  right=0.8mm,
  top=0.8mm,
  bottom=0.8mm,
  listing options={
    basicstyle=\ttfamily\tiny,
    breaklines=true,
    breakautoindent=false,
    breakindent=0pt,
    postbreak={},
    columns=fullflexible,
    keepspaces=true,
    showstringspaces=false
  }
}
Task: For the target segment, determine whether its action or state becomes more specific when earlier segments are taken into account.
Ignore person coreference. Be careful not to over-label.
Mark true only when earlier segments are needed to concretize the target's action, state, object state, posture, or constraint. This includes cases where the target is understandable alone but becomes specifically interpretable only with earlier context.
If the target is a complete standalone action/state description, mark false. Do not mark true for mere temporal continuity or simple repetition when earlier context does not make the target more specific.

Return JSON only with exactly these keys:

{
  "event_anaphora": true or false,
  "depends_on_segment_ids": [],
  "anaphoric_expression_in_target": "",
  "antecedent_expression": "",
  "dependency_type": "event | state | object | none",
  "reason": ""
}

Earlier segments:
{previous_segments}

Target segment:
{target_segment}
\end{tcblisting}
\end{minipage}
\hfill
\begin{minipage}[t]{0.48\textwidth}
\begin{tcblisting}{
  title=Prompts for Episodic Caption Rewriting,
  colback=gray!5,
  colframe=gray!60,
  fonttitle=\bfseries,
  listing only,
  height=0.78\textheight,
  valign=top,
  left=0.8mm,
  right=0.8mm,
  top=0.8mm,
  bottom=0.8mm,
  listing options={
    basicstyle=\ttfamily\tiny,
    breaklines=true,
    breakautoindent=false,
    breakindent=0pt,
    postbreak={},
    columns=fullflexible,
    keepspaces=true,
    showstringspaces=false
  }
}
[Prompt for Part 1]

You rewrite human action captions. Output exactly one sentence in English.

Constraints:
- Must start with "The person".
- Use general verbs.
- Avoid verbs such as looking, watching, seeing, observing, noticing, or speaking when they do not make posture or physical action concrete.
- Must be a single sentence.
- Must be objective and action-focused.
- Remove subjective or interpretive wording.
- Must be 10 to 15 words.
- No extra commentary, no quotes, no lists.

Input caption:
{original_caption}


[Prompt for Part 2 and Later]

You rewrite split-video action captions into a coherent episode narrative.

This is {part_id}.

Previous rewritten captions in chronological order:
{previous_rewritten_captions}
Current part original caption:
{current_original_caption}

Output exactly one sentence in English with these constraints:
- Must start with "The person".
- Use general verbs.
- Avoid verbs such as looking, watching, seeing, observing, noticing, or speaking when they do not make posture or physical action concrete.
- Must be a single sentence.
- Must be 10 to 15 words.
- It should sound natural if someone prepends "Then, " to this caption after the previous captions.
- If a state or constraint introduced in previous captions is still continuing, prefer a natural rephrasing when possible.
- Keep the caption concise while preserving continuity with the previous captions.
- Keep objective action only.
- Remove subjective or interpretive wording.
- No extra commentary, no quotes, no lists.
\end{tcblisting}
\end{minipage}
\subsection{Captioning Model Comparison}
\begin{table}[t]
\centering
\caption{
Comparison of caption annotation methods. BBox denotes the use of bounding-box inputs, and Ep. denotes episodic caption rewriting.
Human Annotator denotes the average score of two human annotators and serves as the reference.
Higher is better for all metrics.
}
\label{tab:caption_annotation_comparison}
\small
\setlength{\tabcolsep}{5.5pt}
\renewcommand{\arraystretch}{1.12}
\begin{tabular}{llcc}
\toprule
Method & Input & BERT $\uparrow$ & ROUGE-L $\uparrow$ \\
\midrule
Human Annotator & Video+BBox & 0.9043 & 0.3167 \\
\midrule
Gemini 2.5 Pro~\citep{gemini25pro} & Video+BBox & \underline{0.8744} & 0.1719 \\
Gemini 2.5 Flash~\citep{gemini25pro} & Video+BBox & 0.8647 & 0.1581 \\
Qwen2.5-VL Instruct~\citep{qwen25vl} & Video+BBox & 0.8655 & \textbf{0.1809} \\
PoseFix~\citep{posefix} & Pose & 0.8064 & 0.0346 \\
PoseScript~\citep{posescript} & Pose & 0.8252 & 0.1303 \\
\midrule
Gemini 2.5 Pro + Episodic & Video+BBox+Ep. & \textbf{0.8758} & \underline{0.1797} \\
\bottomrule
\end{tabular}
\end{table}
To compare annotation quality across different models, we use the annotations provided by human annotators as references and evaluate each method using BERT-score, which measures semantic similarity, and ROUGE-L, which measures string-level similarity.
First, one expert created reference annotations for each segment, and these annotations were then validated by three experts to define the GT annotations.
We then generated annotations for the same segments using different models and compared them against the GT annotations.
The results are shown in Tab.~\ref{tab:caption_annotation_comparison}.

Among models that annotate each segment independently, Gemini 2.5 Pro achieves the highest BERT-score and also performs competitively in ROUGE-L.
When Episodic Annotation is applied with Gemini 2.5 Pro, it achieves the best BERT-score and the second-best ROUGE-L.
These results suggest that Episodic Annotation, which incorporates context from preceding segments, is effective for generating descriptions that are closer to human annotations.

\subsection{Annotation Quality}

\begin{figure*}[t]
\centering

\begin{minipage}[t]{0.48\textwidth}
\vspace{0pt}
\begin{tcblisting}{
  title=User Study Instructions,
  colback=gray!5,
  colframe=gray!60,
  fonttitle=\bfseries,
  listing only,
  breakable=false,
  left=1mm,
  right=1mm,
  top=1mm,
  bottom=1mm,
  listing options={
    basicstyle=\ttfamily\tiny,
    breaklines=true,
    breakautoindent=false,
    breakindent=0pt,
    postbreak={},
    columns=fullflexible,
    keepspaces=true,
    showstringspaces=false
  }
}
In this study, you will evaluate the same target text from two different perspectives.

Each sample corresponds to one 5-second segment from a 30-second human motion sequence, which is divided into six 5-second segments.

First, read the target text describing the 5-second segment. At this point, imagine the human motion as concretely as possible based only on this text.

Next, you will be shown all preceding texts from the same 30-second clip, describing the motions before the target segment. Judge whether these preceding texts clearly add information and help you imagine the target motion more concretely than the target text alone.

- Yes: The preceding context clearly makes the target motion more concrete, such as the action, posture, object, or continuing state.
- No: The target text alone is sufficient to understand the motion, and the preceding context does not substantially change the interpretation.

After that, you will be shown the automatic annotation result. Judge whether the automatic annotation is correct.

- Yes: You think the automatic annotation is correct.
- No: You think the automatic annotation is incorrect.

Then, you will evaluate the same target text again by watching the corresponding video. Judge whether the target text is an appropriate annotation for the human motion shown in the video.

- Yes: The text generally describes the main human motion in the video correctly.
- No: The text clearly differs from the video, incorrectly describes an important motion, or includes content that cannot be confirmed from the video.

Do not judge a sample as context-dependent simply because the same person continues moving over time. Instead, judge whether the preceding context specifically helps clarify or complete the meaning of the target text.

For the video-based evaluation, minor differences in wording or phrasing are acceptable. Base your judgment on whether a person reading the text would largely understand the main motion shown in the video without serious misunderstanding.
\end{tcblisting}
\end{minipage}
\hfill
\begin{minipage}[t]{0.48\textwidth}
\vspace{0pt}
\centering
\includegraphics[width=\linewidth]{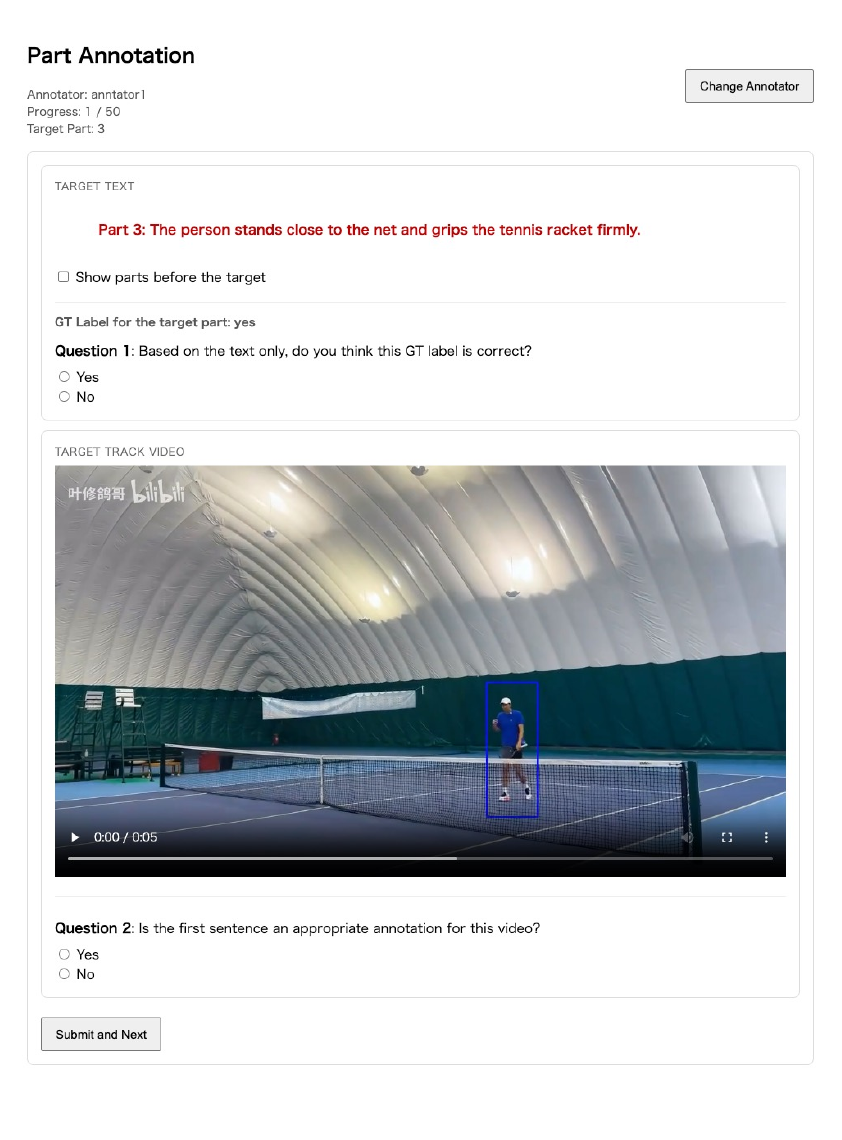}
\end{minipage}

\caption{
User study instructions and app interface.
Left: instructions shown to participants.
Right: screenshot of the app interface used in the user study.
}
\label{fig:userstudyapp}
\end{figure*}
To assess the reliability of our automatic annotations, we conduct two user studies and report both annotation accuracy and inter-annotator agreement measured by Gwet’s AC1.
In each study, we randomly sample 50 motion segments and ask three annotators to provide binary judgments.
Fig.~\ref{fig:userstudyapp} shows the user study instructions and the app interface used for this evaluation.
In the first study, the annotators evaluate whether the automatically assigned anaphora labels are correct.
In the second study, they evaluate whether the generated text annotations are consistent with the corresponding motion segments.
For each segment, we compute the average binary score across annotators as the annotation accuracy, and we use Gwet’s AC1 to quantify inter-annotator agreement.
As a result, the automatically assigned anaphora labels achieve an accuracy of 75\% with Gwet’s AC1 = 0.58, while the generated text annotations achieve an accuracy of 81\% with Gwet’s AC1 = 0.64.

To further investigate the failure modes of automatic annotation, we conduct an additional human evaluation with three annotators on 150 anaphoric segments randomly sampled from the ELMABench test set.
For each sample, the annotators assess whether the annotated dependency can be confirmed in the original video, whether the dependency source segment is correctly identified, and whether the caption contains incorrect action descriptions or left-right/body-part confusion.
The results show that the annotated dependency can be confirmed in the original video for 96\% of the evaluated samples, while the dependency source segment is correctly identified in 78\%.
We additionally adopt a conservative dependency-error criterion, under which a dependency is considered erroneous if either the dependency itself cannot be confirmed in the original video or its source segment is incorrectly identified.
Under this criterion, 22\% of the evaluated samples contain unsupported or incorrectly grounded dependencies, with an inter-annotator agreement of Gwet's AC1 = 0.58.
Among captions judged to contain errors, incorrect action descriptions account for 63\% of the errors, followed by left-right/body-part confusion at 31\%.
These results suggest that automatic annotation can reliably identify the presence of cross-segment dependencies in most cases, while some errors remain in identifying dependency source segments and describing fine-grained motions.
Although automatic annotation cannot completely eliminate such errors, it enables the construction of large-scale datasets with diverse cross-segment dependencies that would be difficult to obtain through manual annotation alone.


\section{Training Details}
\label{app:training_details}
\subsection{Implementation Details} 
To ensure accurate assessment in our new setting, we fine-tuned the evaluation network using the ELMA training set. Specifically, we initialized the evaluator with the \textit{Decomp\_v3 AutoEncoder} pretrained on HumanML3D, and fine-tuned it for text-motion matching with a batch size of 1024 and a learning rate of $1 \times 10^{-4}$ for 37 epochs. To align with the word length distribution of our training data, we also expanded the \texttt{max\_text\_len} parameter to 24.

For the generation evaluation, we strictly use the identical test split for all models. Each model generates a 30-second continuous motion sequence guided by the multi-segment instructions. Importantly, the evaluation metrics are computed exclusively on the target segments that are pre-identified as anaphoric, preventing score influence from easier segments.

During the training phase for both A2M and T2M settings, each baseline model is initialized with its respective HumanML3D-pretrained checkpoint and fine-tuned for an additional 50,000 steps. All experiments were conducted using a single NVIDIA RTX 6000 Ada Generation GPU. 

Following standard protocols, each evaluation is performed 10 times, and we report the mean values along with their 95\% confidence intervals.

The released processed dataset and evaluation code enable reproduction of the main benchmark experiments, including baseline training, motion generation under the T2M and A2M settings, metric computation, and evaluation on the anaphoric subset.
However, fully reproducing the dataset construction pipeline, including video collection, 3D motion recovery, episodic caption generation, and anaphora detection, requires access to the original source videos, which are not publicly redistributed due to privacy and licensing considerations.
Similarly, verifying annotation quality against the original videos requires access to these sources.

\section{Broader Impact}
\label{app:broader_impact}

\subsection{Positive Societal Impact}
ELMA and ELMABench target the setting of long-term, context-dependent text-to-motion
generation, which underlies a wide range of beneficial applications.
Reliable long-term motion generation can reduce the cost of motion-capture pipelines
for indie game developers and animators, enable continuously controllable digital
avatars for virtual and augmented reality, support previsualization in film production,
and provide motion priors for humanoid robotics imitation learning.
By making episodic dependency-rich data publicly available under the CC BY-NC 4.0 license,
ELMA also lowers the barrier to entry for academic groups studying long-term
human motion modeling, an area where dataset access has historically been a
bottleneck.

\subsection{Negative Societal Impact and Mitigations}

\paragraph{Risk of misuse for impersonation.}
Realistic human motion synthesis can, in principle, contribute to deepfake-style content.
We mitigate this in two ways. First, ELMA distributes only SMPL-X body-pose
parameters; identity-revealing body-shape parameters (\texttt{betas}) are
\emph{stripped} from all released sequences, and original videos are not redistributed.
Second, ELMA is released under the CC BY-NC 4.0 license (see Appendix~\ref{app:license}), which prohibits commercial use without separate permission.

\paragraph{Privacy of in-the-wild video subjects.}
Source clips are sampled from publicly accessible online platforms.
We do not redistribute the original videos themselves, nor do we release source video URLs.
Instead, we release only derivative motion parameters without body-shape parameters (\texttt{betas}) and text annotations.
This design reduces the risk of directly identifying or redistributing the original source content.

\paragraph{Demographic and cultural bias.}
Online videos do not reflect a uniform demographic distribution.
Action categories popular in particular geographic regions, age groups, or fitness communities are likely overrepresented, while underrepresented populations may be correspondingly underrepresented.
Models trained on ELMA may inherit these biases and produce more reliable motions for over-represented activity types.
We encourage users to report observed biases and to combine ELMA with demographically curated datasets for downstream applications where fairness is critical.

\paragraph{Surveillance and tracking concerns.}
Although ELMA is intended for generative modeling, motion data and the accompanying collection pipeline (keypoint tracking, mask propagation, mesh recovery) could, in principle, be repurposed for surveillance.
To mitigate these risks, ELMA is released under the CC BY-NC 4.0 license, and we do not redistribute original videos, extracted frames, source-video URLs, or body-shape parameters (\texttt{betas}).
These measures help reduce privacy risks, although they do not eliminate the possibility of misuse.

\paragraph{Annotator welfare.} 
For the user studies in Appendix~\ref{app:annotation}, annotators were recruited internally from the authors' institution. They were compensated at a rate of 1,230 JPY per hour, which meets or exceeds the local minimum wage. The study protocol was reviewed and approved by the authors' institutional ethics review board prior to data collection. All annotators were provided with task instructions, an estimated completion time, and information about the study before participation.

\section{License}
\label{app:license}

ELMA is released as a derivative dataset that combines (i) motion parameters
recovered from publicly accessible online videos using SMPL-X-based mesh recovery
and (ii) text annotations produced by a VLM/LLM-based pipeline.
Because of this composition, ELMA inherits constraints from multiple upstream
licenses, summarized below.
\vspace{-8pt}
\subsection{ELMA Dataset License}
\vspace{-5pt}
ELMA is available on Hugging Face under the \href{https://creativecommons.org/licenses/by-nc/4.0/}{CC BY-NC 4.0} license.
The released dataset contains processed motion parameters, text annotations,
dependency labels, and metadata, but does not redistribute the original source
videos or source-video URLs.
\vspace{-5pt}
\begin{itemize}
    \setlength{\itemsep}{-2.5pt}
    \setlength{\topsep}{-5pt}
    \setlength{\parsep}{0pt}
    \item[$\bullet$] \textbf{Motion parameters (SMPL-X pose sequences).}
    Distributed under the SMPL-X Model License
    (non-commercial scientific research purposes only).
    Body-shape parameters (\texttt{betas}) are removed from all released sequences;
    only joint-pose-related parameters required for motion playback are retained.
    Users must have already accepted the SMPL-X license at \url{https://smpl-x.is.tue.mpg.de/}
    to use these parameters in compliance with the upstream terms.

    \item[$\bullet$] \textbf{Text annotations and dependency labels.}
    Distributed under \href{https://creativecommons.org/licenses/by-nc/4.0/}{CC BY-NC 4.0}.
    This permits academic redistribution and adaptation with attribution, while
    explicitly disallowing commercial use.

    \item[$\bullet$] \textbf{Source code.} Source code is publicly available in the accompanying repository under the license specified therein.
\end{itemize}
\vspace{-8pt}
\subsection{Compliance with Source Video Platforms}
\vspace{-5pt}
Source clips originate from publicly accessible online video platforms.
We reduce redistribution and privacy risks in the following ways:
\vspace{-5pt}
\begin{itemize}
    \setlength{\itemsep}{-2.5pt}
    \setlength{\topsep}{-5pt}
    \setlength{\parsep}{0pt}
    \item[$\bullet$] We do not redistribute the original video files, audio tracks, extracted frames, or source-video URLs.
    \item[$\bullet$] We release only derivative motion parameters without body-shape parameters (\texttt{betas}), together with textual annotations and dependency labels.
    \item[$\bullet$] The released data is intended for non-commercial research use under the dataset license.
    \item[$\bullet$] Users are not permitted to use ELMA for person identification, re-identification, biometric analysis, surveillance, or attempts to recover the original source content.
\end{itemize}
\vspace{-8pt}
\subsection{Third-Party Assets and Their Licenses}
\vspace{-5pt}
The third-party assets used in our pipeline and their respective licenses or terms are summarized in Table~\ref{tab:external_assets}.
\vspace{-8pt}
\begin{table}[H]
\centering
\caption{
External assets referenced or used in this work.
We do not redistribute these third-party datasets, models, or software unless explicitly stated.
Users should obtain each asset from its original source and comply with its respective license or terms.
}
\label{tab:external_assets}
\small
\setlength{\tabcolsep}{3.5pt}
\renewcommand{\arraystretch}{1.0}
\begin{tabular}{lll}
\toprule
Asset & Use in this work & Upstream license / terms \\
\midrule
SMPL-X~\citep{smplx}
  & Motion representation / rendering
  & SMPL-X Model License \\
VPoser~\citep{smplx}
  & PoseFiltering reference manifold
  & Non-commercial research license \\
AMASS~\citep{amass}
  & Reference distribution for filtering thresholds
  & AMASS License \\
HumanML3D~\citep{humanml3d}
  & Representation / pretrained evaluation network
  & MIT License \\
BABEL~\citep{babel}
  & Action categories / query references
  & BABEL License \\
Kinetics-700~\citep{kinetics700}
  & Action query references
  & CC BY 4.0 annotations \\
GVHMR~\citep{gvhmr}
  & World-grounded mesh recovery
  & Non-commercial research license \\
SAM~2~\citep{sam2}
  & Mask propagation
  & Apache License 2.0 \\
YOLO-Pose~\citep{yolo-pose}
  & 2D keypoint detection
  & GPL-3.0 \\
ByteTrack~\citep{bytetrack}
  & Person tracking
  & MIT License \\
PySceneDetect
  & Scene segmentation
  & BSD 3-Clause License \\
CLIP~\citep{clip}
  & Text encoder for baselines
  & MIT License \\
Gemini 2.5 Pro~\citep{gemini25pro}
  & Captioning / MLLM verification
  & Google API Terms of Service \\
GPT-4o~\citep{gpt-4o}
  & Anaphora detection / query expansion
  & OpenAI API Terms of Service \\
\bottomrule
\end{tabular}
\vspace{-5pt}
\end{table}
\vspace{-8pt}
We acknowledge each upstream provider; users of ELMA are required to comply with
all listed third-party licenses in addition to the ELMA license itself.



\clearpage
\end{document}